%% file: paper.tex
\documentclass[10pt,journal,compsoc]{IEEEtran}

\usepackage{url}

\usepackage[nocompress]{cite}
\usepackage{times}
\usepackage{epsfig}
\usepackage{graphicx}
\usepackage{amsmath}
\usepackage{amssymb}
\usepackage{xspace}
\usepackage{rotating}
\usepackage{units}
\usepackage{multirow}
\usepackage{algorithm}
\usepackage{lettrine}
\usepackage{tikz}
\usepackage{amsthm}
\usepackage{xfrac}
\usepackage{listings}
\usepackage{color}
\usepackage{xr}
\usepackage{bbold}
\usepackage{pdfpages}

\def\etal{\emph{et al}.~}

\newcommand{\mypar}[1]{\noindent \textbf{#1}}
\newcommand{\ip}[2]{{#1}^{\top}{#2}}

\def \P    {\mathbb P}
\def \E    {\mathbb E}
\def \V    {\mathbb V}
\def \real {\mathbb R}
\def \Xset {\mathcal X}

\def \x{\mathbf{x}}

\def \m{\mathbf{m}}
\def \y{\mathbf{y}}

\def \Y{\mathbf{Y}}
\def \X{\mathbf{X}}
\def \N{\mathbf{N}}
\def \Z{\mathbf{Z}}

\def \Z{\mathbf{Z}}
\def \U{\mathbf{U}}
\def \u{\mathbf{u}}
\def \1{\mathbf{1}}
\def \I{\mathbf{I}}

\def\tr{\mathsf{tr}}

\def \Pfn {\P_\text{fn}}
\def \Pfp {\P_\text{fp}}

\begin{document}

\title{Memory vectors for similarity search in high-dimensional spaces}

\author{Ahmet~Iscen,
        Teddy~Furon,
        Vincent~Gripon,
        Michael~Rabbat,        
        and~Herv\'e~J\'egou
\IEEEcompsocitemizethanks{\IEEEcompsocthanksitem A. Iscen and T. Furon are with Inria.\protect\\
E-mail: \{ahmet.iscen,teddy.furon\}@inria.fr
\IEEEcompsocthanksitem V. Gripon is with T\'el\'ecom Bretagne.\protect\\
E-mail: vincent.gripon@telecom-bretagne.eu
\IEEEcompsocthanksitem M. Rabbat is with McGill University.\protect\\
E-mail: michael.rabbat@mcgill.ca
\IEEEcompsocthanksitem H. J\'egou is with Facebook AI research.\protect\\
E-mail: rvj@fb.com
}
\thanks{}}

%

\markboth{ }%
{Iscen \MakeLowercase{\textit{et al.}}: Memory vectors for similarity search in high-dimensional spaces}

\IEEEtitleabstractindextext{%
\begin{abstract}
We study an indexing architecture to store and search in a database of high-dimensional vectors from the perspective of statistical signal processing and decision theory. 
This architecture is composed of several memory units, each of which summarizes a fraction of the database by a single representative vector.
The potential similarity of the query to one of the vectors stored in the memory unit is gauged by a simple correlation with the memory unit's representative vector. 
This representative optimizes the test of the following hypothesis: the query is independent from any vector in the memory unit vs. the query is a simple perturbation of one of the stored vectors.

Compared to exhaustive search, our approach finds the most similar database vectors significantly faster without a noticeable reduction in search quality.
Interestingly, the reduction of complexity is provably better in high-dimensional spaces. 
We empirically demonstrate its practical interest in a large-scale image search scenario with off-the-shelf state-of-the-art descriptors.
\end{abstract}

}

\maketitle

\IEEEpeerreviewmaketitle

\input{intro.tex}
\input{prob.tex}

\input{comu.tex}
\input{coma.tex}

\input{applications.tex}

\input{appendix.tex}

\section{Conclusion}
In this paper, we take a statistical signal processing point of view for image indexing, instead of traditional geometrical approaches in the literature. This shift of paradigm allows us to bring theoretical justifications for representing a set of vectors.
We have presented and analyzed two strategies for designing memory vectors, enabling efficient membership tests for real-valued vectors. 
We have also showed two possible assignment strategies and analyzed their performance theoretically and experimentally.
For random assignment, the optimized \textit{pinv} construction gives better results than the simple \textit{sum} aggregation.
On the other hand, when the vectors in the same memory unit share some correlation, \textit{sum} is on par with the  \textit{pinv} construction as for the quality of the hypothesis test.
Yet, the \textit{pinv} construction when used in the weakly supervised assignment offers a lower imbalance factor.
This yields less variability of the search runtime from one query to another.
This procedure is done offline and its complexity is often ignored in the image search literature.
On the contrary, we did pay attention to this bottleneck: we proposed to run the weakly supervised assignment by batch and showed that it does not spoil the overall performance of the image search.

\section*{Acknowledgments}
This work was supported, in part, by the Natural Sciences and Engineering Research Council of Canada through grants RGPAS 429296-12 and RGPIN/341596-12, by the CHIST-ERA project ID\_IOT (ANR-16-CHR2-0005), and by the European Research Council under the European Union's Seventh Framework Programme ( FP7 / 2007 - 2013 ) / ERC grant agreement 290901.


%
%

\ifCLASSOPTIONcaptionsoff
  \newpage
\fi

{
\bibliographystyle{IEEEtran}
\bibliography{egbib,egbibJ,teddy}
}

\begin{IEEEbiography}
[{\includegraphics[width=1in,height=1in]{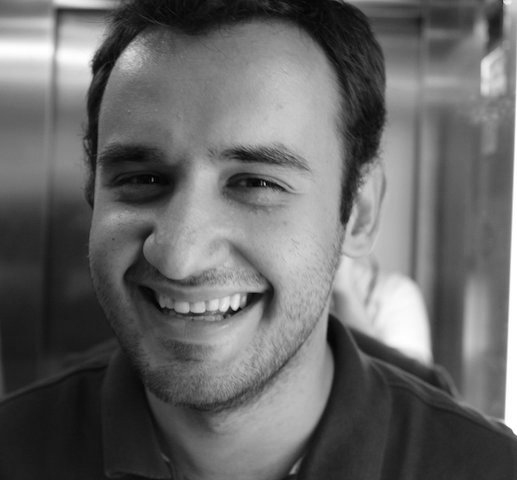}}]{Ahmet Iscen} received his B.Sc. degree from The
State University of New York at Binghamton, and
M.Sc. from Bilkent University. He is a PhD student
at Inria Rennes and University of Rennes I, working
on large-scale image retrieval.
\end{IEEEbiography}

\begin{IEEEbiography}
[{\includegraphics[width=1in,height=1.2in]{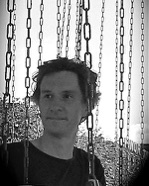}}]{Teddy Furon} is an INRIA senior researcher in the LinkMedia team. His fields of interest are digital watermarking, traitor tracing, security of Content Based Image Retrieval and more broadly security issues related to multimedia and signal processing. He received the DEA degree (M.Sc.) in digital communication in 1998 and the PhD degree in signal and image processing in 2002, both from Telecom ParisTech. He has worked both in industry (Thomson, Technicolor) and in academia (Univ. Cath. de Louvain, Belgium, and Inria Rennes, France). He has been a consultant for the Hollywood MovieLabs competence center. He received the Brittany Best Young Researcher prize in 2006. He is the co-author of 60 conference papers, 18 journal articles, 4 book chapters and 9 patents. He has been associate editor for 4 scientific journals (among them IEEE Trans. on Information Forensics and Security). He co-founded of the startup company LAMARK which protects pictures of independent photographers and photo agencies.
\end{IEEEbiography}

\begin{IEEEbiography}
[{\includegraphics[width=1in,height=1.2in]{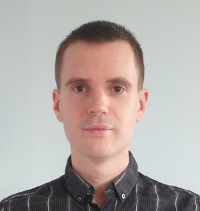}}]{Vincent Gripon} (S'10, M'12) is a permanent researcher with T\'el\'ecom Bretagne (Institut Mines-T\'el\'ecom), Brest, France. He obtained his M.S. from \'Ecole Normale Sup\'erieure of Cachan and his Ph.D. from T\'el\'ecom Bretagne. His research interests lie at the intersection of information theory, computer science and neural networks. He is the co-creator and organizer of an online programming contest named TaupIC which targets French top undergraduate students.
\end{IEEEbiography}

\begin{IEEEbiography}
[{\includegraphics[width=1in,height=1.25in]{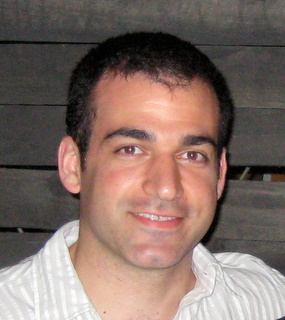}}]{Michael Rabbat} received the B.Sc. degree from the University of Illinois, Urbana-Champaign, in 2001, the M.Sc. degree from Rice University, Houston, TX, in 2003, and the Ph.D. degree from the University of Wisconsin, Madison, in 2006, all in electrical engineering. He joined McGill University, Montréal, QC, Canada, in 2007, and he is currently an Associate Professor and William Dawson Scholar. During the 2013–2014 academic year he held visiting positions at Télécom Bretegne, Brest, France, the Inria Bretagne-Atlantique Reserch Centre, Rennes, France, and KTH Royal Institute of Technology, Stockholm, Sweden. He was a Visiting Researcher at Applied Signal Technology, Inc., Sunnyvale, USA, during the summer of 2003. He co-authored the paper which received the Best Paper Award (Signal Processing and Information Theory Track) at the 2010 IEEE International Conference on Distributed Computing in Sensor Systems (DCOSS), and his work received an Honorable Mention for Outstanding Paper Award at the 2006 Conference on Neural Information Processing Systems (NIPS). He currently serves as Senior Area Editor for \textsc{IEEE Signal Processing Letters}, and as Associate Editor for \textsc{IEEE Transactions on Signal and Information Processing over Networks} and \textsc{IEEE Transactions on Control of Network Systems}. His research interests include statistical signal processing, graph signal processing, distributed algorithms for optimization and inference, gossip and consensus algorithms, and network modelling and analysis.
\end{IEEEbiography}

\begin{IEEEbiography}[{\includegraphics[width=1in,height=1.2in,clip,keepaspectratio]{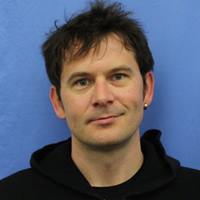}}]{Herv\'e J\'egou}is a research manager at Facebook AI research. He holds a PhD (2005) from University of Rennes defended on the subject of error-resilient compression and joint source channel coding. He joined Inria in 2006 as a researcher, where he turned to Computer Vision and Pattern Recognition. He joined Facebook AI Research in 2015. 
\end{IEEEbiography}


\end{document}

%% file: intro.tex

\IEEEraisesectionheading{\section{Introduction}}
\label{sec:intro}

\IEEEPARstart{W}{e} consider the problem of searching for vectors similar to a query vector in a large database.
The typical applications are the search and exploration in Big Media Data where documents are represented by feature vectors~\cite{Wang:2015fe}.
In this context, many papers report how the curse of dimensionality (due to the size of the vectors) makes indexing techniques ineffective~\cite{WSB98,ML14}. The recent paper~\cite{ML14} describing and analyzing the popular FLANN method experimentally  observes that even this state-of-the-art method performs poorly on synthetic high-dimensional vectors, and the authors conclude that ``random datasets are one of the most difficult problems for nearest neighbor search''.

Some strategies have been proposed to (partly) overcome this problem. For instance, the vector approximation file~\cite{WSB98} first relies on exhaustive search with approximate measurements and then computes the exact similarities only for a subset of vectors deemed of interest. The cosine sketch~\cite{C02} approximates cosine similarity with faster Hamming distance. Other works like spectral hashing~\cite{WTF09}, Euclidean sketches~\cite{DCL08}, product quantization~\cite{JDS11} and inverted multi-index~\cite{BL12} also rely on compact codes to speed up neighbor search while compressing the data.
An interesting strategy is the Set Compression Tree~\cite{AZ14}, which uses a structure similar to a k-d tree to compress a set of vectors into an extremely compact representation.
Again, this method is dedicated to small dimensional vectors (its authors recommend the dimension be smaller than $\log_{2}(N)$ where $N$ is the size of the database) so that it is used in conjunction with a drastic dimension reduction via PCA to work with classical computer vision descriptors.

The main contribution of the paper is a similarity search approach specifically adapted to high-dimensional vectors
such as those recently introduced in computer vision to represent images~\cite{PD07,JDSP10}. 
The proposed indexing architecture consists of memory units, each of which is associated with several database vectors. 
A representative, called a memory vector, is produced for each memory unit and defined such that one can quickly and reliably infer whether or not at least one similar vector is stored in this unit by computing a single inner product with a new query. 

Our approach is similar to the descriptor pooling problem in computer vision, but at a higher level. Many successful descriptors, such as BOV~\cite{CDFWB04,SZ03}, VLAD~\cite{JDSP10}, FV~\cite{PD07}, and EMK~\cite{BS09}, encode and aggregate a set of local features into global representations. Yet, the new representation has a larger dimension than the local features.
We use a similar approach at a higher-level: instead of aggregating local features into one global image representation, we aggregate global representations into group representations, so called memory vectors. These have no semantical meaning. Their purpose is to allow efficient search. Another difference is that we keep the same ambient dimension.

A second contribution of the paper is to study similarity search from the perspective of statistical signal processing and decision theory. Our analysis provides insight into when and why the proposed approach provides low complexity (fast response time) without sacrificing accuracy. A third contribution is the experimental work using computer vision Big Media datasets as large as 100 million images. Our results suggest that the approach we propose can achieve accuracy comparable to state-of-the-art while providing results 5--10$\times$ faster.

The paper is organized is as follows. Section~\ref{sec:prob} gives a formal problem statement. Section~\ref{sec:comu} focuses on the design of a single memory vector. We formalize the similarity of a query  with the vectors of one memory unit as a hypothesis test. We derive the optimal representative vector under some design constraints and show how to compute it in an online manner.
Section~\ref{sec:coma} proposes and analyzes two different ways to assign database vectors to memory units: random assignment and weakly supervised assignment which packs similar vectors into memory units.  
We provide a theoretical and experimental analysis of the different design and assignment strategies.
Section~\ref{sec:app} presents the results of experiments that evaluate our approach on standard benchmarks for image search.
We use descriptors (vectors describing images) extracted with the most recent state-of-the-art algorithm in computer vision~\cite{JZ14}. Our results show the potential of our approach for this application.
Finally, we present the complete analysis of our method in Section~\ref{sec:analysis}.

%% file: prob.tex

\section{Problem Formulation}
\label{sec:prob}

Let $\Xset = \{\x_1,\dots,\x_n\}$ denote a collection of $d$-dimensional vectors. We assume that all vectors are normalized so that $\| \x_i \|_2^2 = 1$ for all $i=1,\dots,n$.

Given a query vector $\y \in \real^d$ with $\|\y\|_2^2 = 1$, our objective is to determine which vectors $\x_i \in \Xset$ are most similar to $\y$. Determining which images would be ``most similar'' to a human is subjective and not easily quantified. The real image datasets used in the experiments we report in Section~\ref{sec:app} include a set of queries and the corresponding human-determined response sets, which are treated as ground truth for those experiments. While humans can provide result sets for individual queries, generalizing to produce result sets for never-seen-before queries is still an open problem.

More generally, when such a ground truth is not given to measure performance, since all vectors lie on the unit sphere, we measure similarity using the inner product $\ip{\x_i}{\y}$. Then the problem of determining the most similar vectors to the query can be stated more precisely as: given $\alpha_0 > 0$, find all vectors $\x_i \in \Xset$ such that $\ip{\x_i}{\y} \ge \alpha_0$. The accuracy of a technique can be measured in terms of its precision and recall.

Clearly a naive baseline approach to this problem would compute all $n$ inner-products $\ip{\x_i}{\y}$. Although it may provide perfect accuracy, this approach would have computational complexity (strongly related to run-time) of $\mathcal{O}(nd)$ operations. This is generally unacceptable when $n$ and/or $d$ are large, and we aim to obtain high accuracy while reducing the complexity.

%% file: comu.tex

\section{Memory vectors}
\label{sec:comu}

Given a memory unit $\Xset = \{\x_1,\dots,\x_n\} \subset \real^d$, our objective in this section is to produce a representative, a so-called memory vector, such that, given a query vector $\Y$ regarded as a random variable, we can efficiently perform a ``similarity'' test answering:
\emph{is $\Y$ a quasi-copy of, or similar to, at least one of the vectors of the memory unit?}

For the sake of analysis, this section assumes that the vectors $\x_i$ are independent and identically distributed samples from a uniform distribution on the $d$-dimensional unit hypersphere. 
We model the query as a random vector $\Y$ distributed according to one of the two laws:
\begin{itemize}
\item Hypothesis $\mathcal{H}_{0}$: $\Y$ is not related to any vector in $\Xset$. $\Y$ is then uniformly distributed on the unit hypersphere.
\item Hypothesis $\mathcal{H}_{1}$: $\Y$ is related to one vector in $\Xset$, say $\x_{1}$ without loss of generality. We write this relationship as $\Y=\alpha\x_{1}+\beta \Z$, where $\Z$ is a random vector orthogonal to $\x_{1}$ and $\|\Z\|=1$. This means that $\Y$ is more similar to $\x_{1}$ as $\alpha$ gets closer to 1. We have $\alpha^{2}+\beta^{2}=1$ because $\|\Y\|=1$.
\end{itemize}

We look for a representation scheme satisfying the following design constraints. 
First, the set of vectors $\Xset$ is summarized by a single vector of the same dimension, $\m(\Xset)  \in \real^d$, called the \emph{memory vector} and denoted by $\m$ when not ambiguous. Second, the potential membership of query $\Y$ to $\Xset$ is tested by thresholding the inner product $\ip{\m}{\Y}$.

\subsection{Sum-memory vector: Analysis}
\label{sec:sum}

A very simple way to define the memory vector is
\begin{equation}
\m(\Xset) = \sum_{\x\in\Xset} \x, 
\end{equation}
where we assume that $\Xset$ is composed of $n$ different vectors.
Albeit naive, this strategy offers some insights when considering high-dimensional spaces.

Section~\ref{sec:pdf} derives the pdf of the score $\ip{\m}{\Y}$ under $\mathcal{H}_{0}$ when $\m$ is a known vector. This score has expectation 0 and variance $\|\m\|^{2}/d$, and it is asymptotically distributed as $\mathcal{N}(0,\|\m\|^{2}/d)$ as $d\rightarrow\infty$.
This gives an approximate pdf of $\ip{\m}{\Y}$ under $\mathcal{H}_{0}$.
In contrast, under $\mathcal{H}_{1}$, the inner product equals
\begin{equation}
\label{eq:H1Naive}
\ip{\m}{\Y} = \alpha + \alpha\ip{\m(\Xset^{\prime})}{\x_{1}}+ \beta\ip{\m(\Xset^{\prime})}{\Z},
\end{equation}
with $\Xset^{\prime}=\Xset-\{\x_{1}\}$. This shows two sources of randomness: the interference of $\x_{1}$ with the other vectors in $\Xset$ and with the noise vector $\Z$. 
Assuming that $\Y$ is statistically independent of the vectors in $\Xset^{\prime}$ (this implies that the vectors of $\Xset$ are mutually independent), we have
\begin{eqnarray}
\E_{\Y}[\ip{\m}{\Y} |\mathcal{H}_{1}]&=&\alpha,\nonumber\\
\V[\ip{\m}{\Y} | \mathcal{H}_{1}]&=&\|\m(\Xset^{\prime})\|^{2}/d.
\end{eqnarray}

Assuming that $\Xset$ is composed of $n<d$ statistically independent vectors on the unit hypersphere also gives $\E_{\Xset}[\|\m(\Xset)\|^{2}]=n$ and $\E_{\Xset^{\prime}}[\|\m(\Xset^{\prime})\|^{2}]=n-1$.
To summarize, for large $d$, we expect the following distributions:
\begin{eqnarray}
\mathcal{H}_{0}:& \ip{\m}{\Y}\sim\mathcal{N}(0,n/d),\label{eq:ModelH0Sum}\\
\mathcal{H}_{1}:& \ip{\m}{\Y}\sim\mathcal{N}(\alpha,(n-1)/d).\label{eq:ModelH1Sum} 
\end{eqnarray}
Making a hard decision by comparing the inner product to a threshold $\tau$, the error probabilities (false positive and false negative rates) are given by
\begin{align}
\label{eq:sumfpfn}
\Pfp \approx & 1-\Phi\left(\tau\sqrt{d/n}\right) \\
\Pfn \approx & \Phi\left((\tau-\alpha)\sqrt{d/(n-1)}\right),
\end{align}
where $\Phi(x)= \frac{1}{\sqrt{2\pi}}\displaystyle\int^x_{-\infty}\!e^{\sfrac{-t^2}{2}}dt$. 

The number of elements one can store in a sum-memory vector is linear with the dimension of the space when vectors are drawn uniformly on the unit hypersphere. This construction is therefore useful for high-dimensional vectors only, as opposed to traditional indexing techniques that work best in low-dimensional spaces.  

If the vectors were pair-wise orthogonal, 
the dominant source of randomness (the interference between $\x_{1}$ and vectors of $\Xset'$) is cancelled in~\eqref{eq:H1Naive}. The variance under $\mathcal{H}_{1}$ reduces to $\beta^{2}(n-1)/d$. 
This prevents any false negative if $\beta\rightarrow0$.
We further exploit this intuition that orthogonality helps in the next section.

\subsection{Optimization of the hypothesis test per unit}
\label{sec:pinv}

We next consider optimizing the construction of the memory vector of a given set $\Xset$. 
Denote the $d\times n$ matrix $\X=[\x_{1},\ldots,\x_{n}]$. We impose that, for all $i$, $\ip{\x_{i}}{\m(\Xset)}=1$ exactly and not only in expectation, as assumed above. In other words, $\X^{\top}\m=\1_{n}$ where $\1_{n}$ is the length-$n$ vector with all entries equal to 1. Achieving this, when $\Y=\x_{1}$, we eliminate the interference with the remaining vectors in $\Xset^{\prime}$ which was previously the dominant source of noise. In other words, under $\mathcal{H}_{1}$, Eq.~\eqref{eq:H1Naive} becomes
\begin{equation}
\ip{\m}{\Y} = \alpha + \beta\ip{\m}{\Z}.
\end{equation}

Under $\mathcal{H}_{0}$, the variance of the score remains $\|\m\|^{2}/d$. Therefore, the norm of the memory vector is the key quantity determining the false positive probability. 

We thus seek the representation $\m$ minimizing the energy $\|\m\|^{2}$ subject to the constraint that $\X^{\top}\m=\1_{n}$.
If multiple solutions exist, the minimal norm solution is given by the \textit{Moore-Penrose pseudo-inverse}~\cite{RM71}:  
\begin{equation}
\m^{\star}=(\X^{+})^{\top}\1_{n}.
\label{equ:pinv}
\end{equation}
Since $n<d$, $\m^{\star} = \X(\X^{\top}\X)^{-1}\1_{n}$.
If no solution exists, $\m^{\star}$ is a minimizer of $\|\X^{\top}\m-\1_{n}\|$. 
This formulation amounts to treating the representation of a memory unit as a linear regression problem~\cite{B06} with the objective of minimizing over $\m$ the quantity $\|\X^{\top}\m-\1_{n}\|^{2}$.
Taking the gradient, setting it equal to zero, and solving for $\m$ gives back $\m^{\star}$. 
When possible, and for large $d$, this new construction leads to the distributions:
\begin{eqnarray}
\mathcal{H}_{0}:& \ip{\m^{\star}}{\Y}\sim\mathcal{N}(0,\|\m^{\star}\|^{2}/d)\\
\mathcal{H}_{1}:& \ip{\m^{\star}}{\Y}\sim\mathcal{N}(\alpha,\beta^{2}\|\m^{\star}\|^{2}/d) .
\end{eqnarray} 
The major improvement comes from the reduction of the variance under $\mathcal{H}_{1}$ for small values of $\beta^{2}$, i.e., $\alpha\lesssim 1$. Section~\ref{sec:AppExp} shows that if the vectors of $\Xset$ are uniformly distributed then $\|\m^{\star}\|^{2}$ is larger in expectation than the square norm of the naive sum representation from Section~\ref{sec:sum}. The reduction of the variance under $\mathcal{H}_{1}$ comes at the price of an increase of the variance under $\mathcal{H}_{0}$. However, this increase is small if $n/d$ remains small. For large $d$, we have
\begin{align}
\label{eq:pinvfpfn}
\Pfp \approx & 1-\Phi\left(\tau\sqrt{\frac{d}{n}-1}\right), \\
\Pfn \approx & \Phi\left(\frac{\tau-\alpha}{\beta}\sqrt{\frac{d}{n}-1}\right).
\end{align}

Note that $\beta=\sqrt{1-\alpha^{2}}$ is a decreasing function of $\alpha$. Therefore, if $\tau<\alpha$, $\Pfn$ is a decreasing function of $\alpha$. In particular $\Pfn \rightarrow 0 $ when $\alpha \rightarrow 1$ as claimed above. In contrast to the naive sum approach from Section~\ref{sec:sum}, there is no longer false negative when the query $\Y$ is exactly one of the vectors in $\Xset$. This holds for any value of $\tau<1$ when $\alpha=1$, so that the false positive rate can be as low as $1-\Phi(\sqrt{d/n-1})$.

\noindent{\bf Remark.} This solution is identical (up to a regularization) to the ``generalized max-pooling'' method introduced to aggregate local image descriptors~\cite{MP14}. However in our case the aggregation is performed on the database side only. Our solution is moreover theoretically grounded by a hypothesis test interpretation.

\subsection{Weakly supervised assignment}
\label{sec:kmeansComu}
We now analyze a scenario where the vectors packed in the same memory unit are random but no longer uniformly distributed over the hypersphere: There is some correlation among them.
The vectors in a memory unit are now uniformly distributed over a spherical cap (see Sections~\ref{sec:ModelingKmeanSum} and~\ref{sec:ModelingKmeanPinv}).  
This models the effect of a pre-processing which analyses the database vectors in order to assign similar vectors to memory units.
For instance, Section~\ref{sub:Comparison} uses the $k$-means algorithm to process batches of database vectors.

We derive the same analysis as in the previous subsection with expectations and variances which now depend on the angle of the spherical cap. These expressions are complex and their derivation is detailed in Section~\ref{sec:analysis}.
In summary, the Kullback-Leibler distance between the distributions of $\m^{\top}\Y$ under both hypotheses increases as the spherical cap gets narrow.
In other words, identifying the positive memory units becomes easier when we assign correlated vectors to the same memory unit. Interestingly, this mechanism helps the \textit{sum} construction more than \textit{pinv}, so that when the vectors are very correlated, both constructions indeed perform equivalently.

%% file: coma.tex

\section{Experimental Investigation}
\label{sec:coma}

\begin{figure}
        {\includegraphics[width=0.48\textwidth]{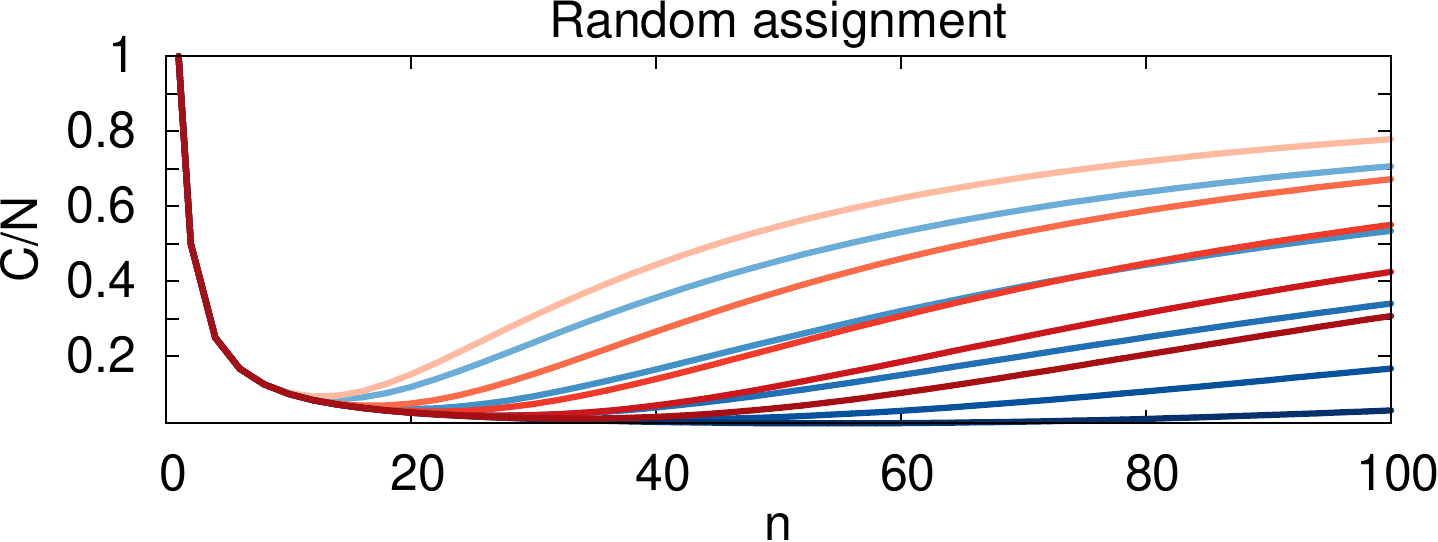}}
    \hfill            
\caption{Ratio of the global cost by the cost of the exhaustive search $C_{\mathcal{H}_{0}}/N$ as a function of $n$ using random assignment on synthetic data. Setup: $d=1000$, $\epsilon = 10^{-2}$. The different curves correspond to values of $\alpha_0 \in\{0.5, 0.6 , 0.7,0.8, 0.9\}$. Red and blue lines correspond to \textit{sum} and \textit{pinv} constructions respectively. Darker shades correspond to higher $\alpha_0$. 
\label{fig:diffMRand}}
\end{figure}

We next consider application scenarios where we need to store a large number~$N$ of vectors and perform similarity search. One memory vector is not sufficient to achieve a reliable test. We therefore consider an architecture that consists of $M$ memory vectors. The search strategy is as follows. A given query vector is compared with all the memory vectors. Then we compare the query with the vectors stored in the memory units associated with the high responses, i.e., those likely to contain a similar vector.

\subsection{Experimental Setup}
\label{sub:ExpSetup}
Our experimental investigations are carried out using synthetic data (vectors uniformly distributed over the hypersphere) as well as real data, which are described in this section.

\mypar{Datasets.}
We use the Inria Holidays~\cite{JDS08}, Oxford5k~\cite{PCISZ07}, and UKB~\cite{NS06} image datasets in our experiments. 
Additionally, we conduct large scale experiments in Holidays+Flickr1M, which is created by adding images from the Flickr1M~\cite{PCISZ07} dataset to the Holidays dataset. Similarly we use the recently introduced Yahoo100M dataset~\cite{TSF+15} to increase the dataset size. 
\smallskip

\mypar{Descriptors.}
We use the state-of-the-art triangular embedding descriptor~\cite{JZ14}, denoted by $\phi_{\Delta}$. We use the off-the-shelf reference implementation provided by the authors, which can be found online.\footnote{\url{http://www.tinyurl.com/democratic-kernel}} Each image is represented by a feature vector.
The only difference is that we do not apply the ``powerlaw normalization'' to better illustrate the benefit of the \emph{pinv} technique for the memory vector construction compared to the \emph{sum} (when applying the powerlaw normalization, both designs perform equally well since the vectors are nearly orthogonal). Ultimately, we have $d=8064$ (or $d=1920$) dimensional feature vectors for each image, obtained by using a vocabulary of size $64$ (resp.~16). For large experiments in Yahoo100M, we use $d=1024$ VLAD descriptors as extracted by~\cite{SPKTV14}.

We also experiment using deep learning features ($d=4096$) provided by Babenko \etal\cite{BSCL14}. As explained in their paper, the performance for the UKB dataset drops with adapted features trained on the Landmarks dataset. Therefore, we use the original neural codes trained on ILSVRC for the UKB dataset, and the adapted features for Holidays and Oxford5k. 

\label{sub:Comparison}
\begin{figure*}
		 \includegraphics[width=0.45\textwidth]{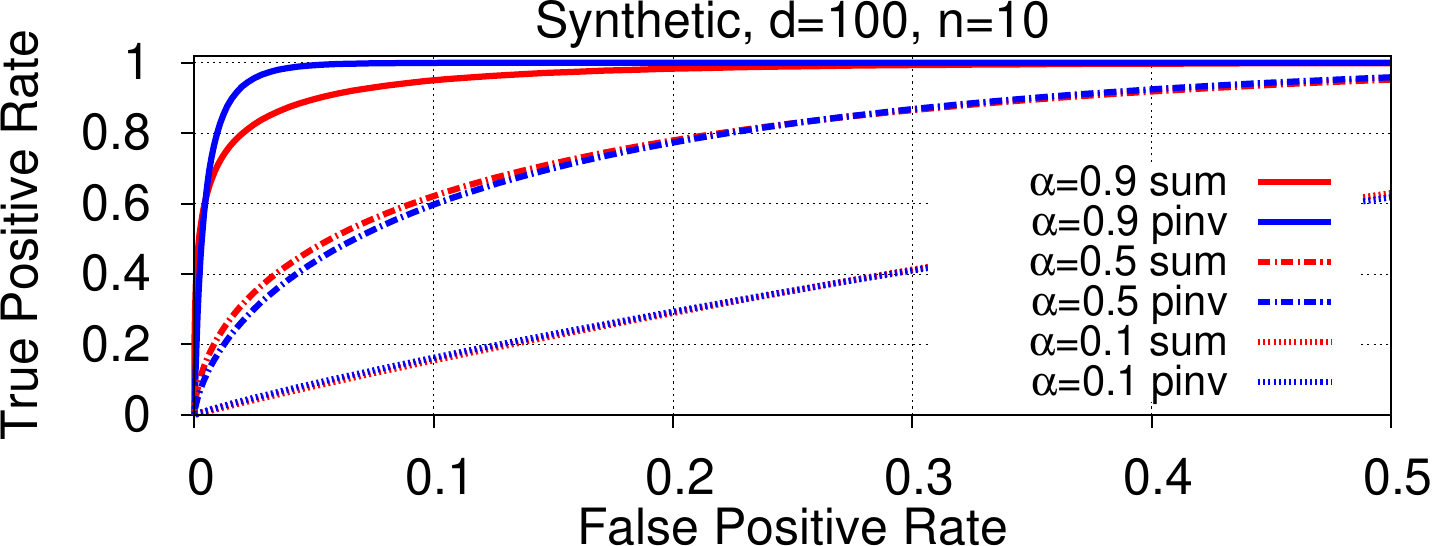}
		 \hfill
        \includegraphics[width=0.45\textwidth]{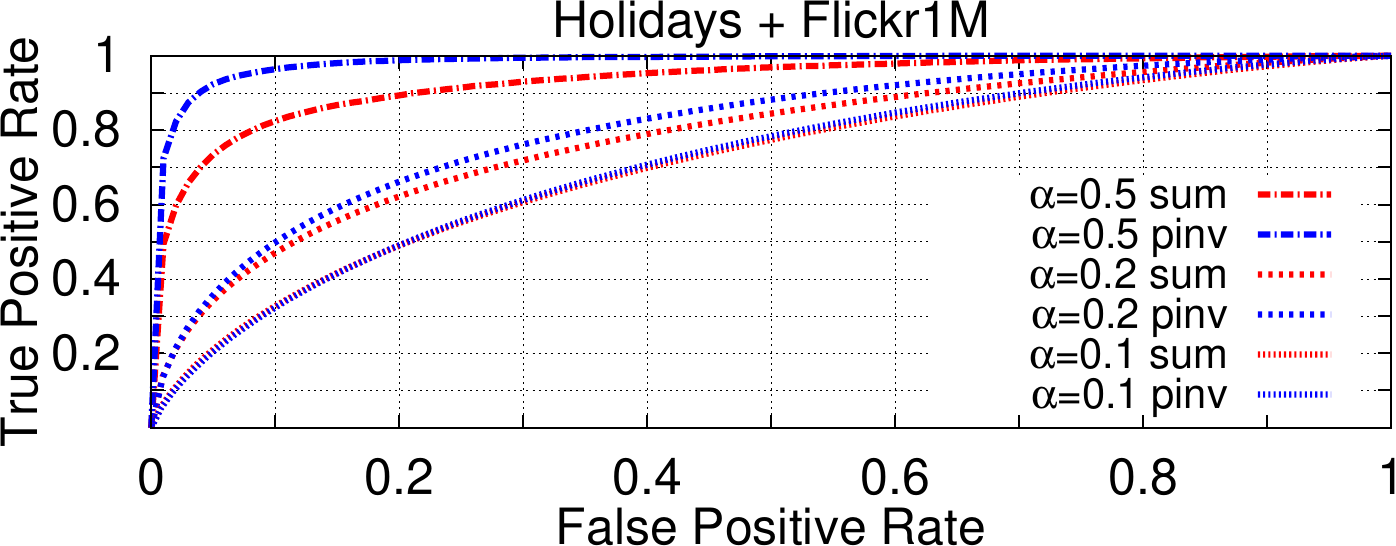}
\caption{ROC curves ($1-\Pfn$ as a function of $\Pfp$) for \textit{sum} and \textit{pinv} constructions evaluated using (\textit{left}) synthetic data with $d=100$ and $n = 10$, (\textit{right}) Holidays+Flickr1M real dataset and $\phi_{\Delta}$ features with $d=1920$ and $n=10$.
\label{fig:retFlickr1Mcos}}
\vspace{-5pt}
\end{figure*}

\subsection{Random assignment}
We suppose that the $N$ vectors in the database are \emph{randomly} grouped into $M$ units of $n$ vectors: $N = nM$. We aim at finding the best value for $n$. When the query is related to the database (i.e., under $\mathcal{H}_1$), we make the following assumption: $\alpha_{0}<\alpha<1$, and we fix the following requirement: $\Pfn<\epsilon<1/2$. Since $\Pfn$ is a decreasing function of $\alpha$, we need to ensure that $\Pfn(\alpha_{0})=\epsilon$. This gives us the threshold $\tau$:
\begin{equation}
\tau = \mu_{\mathcal{H}_{1}}+\sigma_{\mathcal{H}_{1}}\Phi^{-1}(\epsilon),
\end{equation}
with $\mu_{\mathcal{H}_{1}}$ and $\sigma_{\mathcal{H}_{1}}$ being the expectation and the standard deviation of $\ip{\m_{j}}{\Y}$ under $\mathcal{H}_{1}$.
Note that $\Phi^{-1}(\epsilon)<0$ because $\epsilon<1/2$ so that $\tau < \alpha$.
The probability of false positive equals $1-\Phi(\tau/\sigma_{\mathcal{H}_{0}})$ which depends on $n$, denoted by $\Pfp(n)$. This is indeed an increasing function for both memory vector constructions.
Now, we decide to minimize the expectation of the total computational cost $C_{\mathcal{H}_{0}}$ when the query is not related. We need to compute one inner product $\ip{\m_{j}}{\y}$ per unit, and then to compute $n$ inner products $\ip{\x_{i}}{\y}$ for the units giving a positive detection.
In expectation, there are $M \cdot \Pfp(n)$ such units, and so
\begin{equation}
\label{eq:cost}
C_{\mathcal{H}_{0}} = M + M \cdot \Pfp(n) \cdot n=N(n^{-1} + \Pfp(n)).
\end{equation}
The total cost is the sum of a decreasing function ($n^{-1}$) and an increasing function ($\Pfp(n)$).

For the random assignment strategy, there is a tradeoff between having a few big units ($n$ large) and many small units ($n$ small). Fig.~\ref{fig:diffMRand} illustrates this tradeoff for different values of $\alpha_0$ with synthetic data.
It is not possible to find a closed form expression for the cost minimizer $n^{\star}$. 
When $\alpha_{0}$ is close to 1, the threshold is set to a high value, producing reliable tests, and we can pack many vectors into each unit: $n^{\star}$ is large allowing a huge reduction in complexity.
Even when $\alpha_0$ is as small as 0.5, $n^{\star}$ is small but the improvement remains significant.
In the setup of Fig.~\ref{fig:diffMRand}, the proposed approach has a complexity that is less than one tenth of that of searching through all database vectors (equivalent to $n=1$). 
However, in order to increase the efficiency, we introduce an additional $\mathcal{O}(Md) = \mathcal{O}(dN/n)$ memory overhead for storing memory vectors.

Figure~\ref{fig:retFlickr1Mcos} depicts the theoretical and empirical Receiver Operating Characteristic (ROC) curves for different values of $\alpha$. For the synthetic data, $\Pfn$ and $\Pfp$ are evaluated using Eq.~\eqref{eq:sumfpfn} and ~\eqref{eq:pinvfpfn}. As expected the test performs better when $\alpha$ is closer to 1 and when $n \ll d$. 
For the real data, 
we use cosine similarity-based ground truth, since it is directly related with the model we considered theoretically. For each query vector, we deem a database vector as relevant if their cosine similarity is greater than $\alpha_{0}$. To have enough ground-truth vectors, we look for these matching vectors on the Holidays+Flickr1M dataset using various $\alpha_{0}$ values.
This experiment using real data confirms the findings of the theoretical analysis.
The \emph{pinv} construction performs better than the \emph{sum} as long as $\alpha_{0}$ is big as explained in Sect.~\ref{sec:pinv}. 
However, the theoretical analysis is unable to predict performance levels on real dataset. It seems that the vectors of this real dataset have a much lower intrinsic dimensionality than their representational dimension $d=1920$.

\subsection{Weakly supervised assignment}
\label{Sub:COMAWeakly}

A well-known technique in the approximate search literature is to partition the space $\mathbb{R}^{d}$ by clustering the database vectors. This assigns similar database vectors to the same cell~\cite{ML14}.
In Section~\ref{sec:kmeansComu}, we explain the advantage of a weakly supervised assignment by showing that
the distance between the distributions of positive and negative memory units similarities increases.
To show this point experimentally, we modify the spherical $k$-means clustering~\cite{DM01}, so that the clusters are represented using $\emph{pinv}$ (or $\emph{sum}$) in the update stage~\footnote{Note that, when we use such assignment techniques in spherical k-means, the number of vectors per clusters is not evenly distributed. The dot product may be dominated by long cluster representation vectors. Hence, we also propose a normalized version of the assignment, where the cluster representation vectors obtained are normalized to the unit norm.}.

\begin{figure*}
\begin{center}
	\includegraphics[height=2.1cm,trim=0cm 0cm 0cm 0cm,clip]{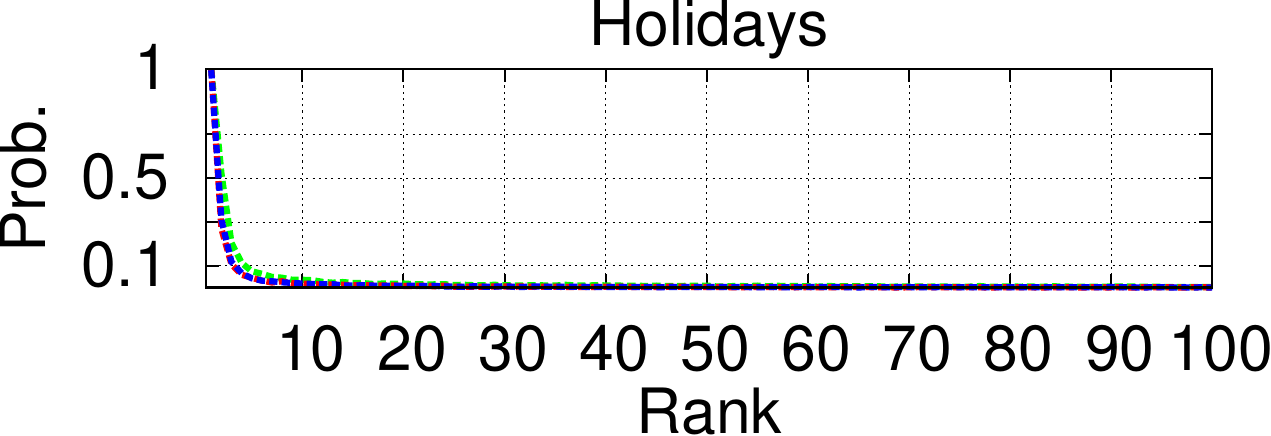}
	\hfill
	\includegraphics[height=2.1cm,trim=1.3cm 0cm 0cm 0cm,clip]{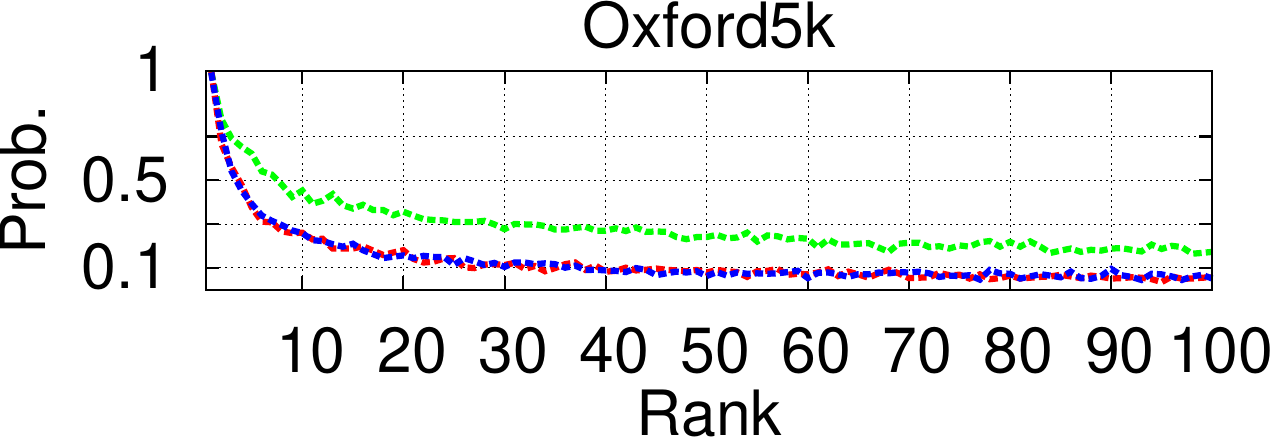}
	\hfill
	\includegraphics[height=2.1cm,trim=1.3cm 0cm 0cm 0cm,clip]{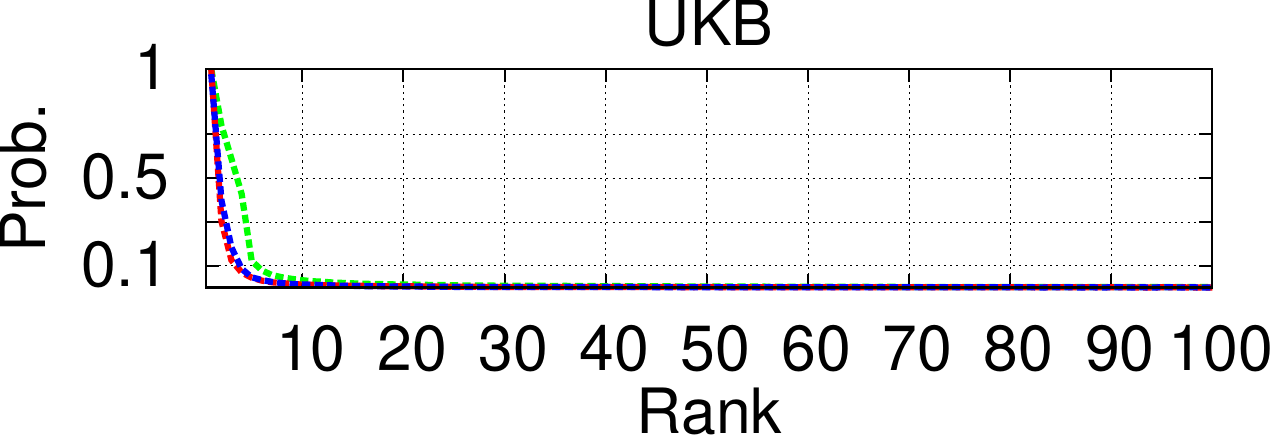}
\end{center}
\caption{Probability that a memory unit contains at least one match with respect to their rank. Green, red and blue lines correspond to random, \textit{sum} spherical k-means and \textit{pinv} spherical k-means respectively. There is a high probability of retrieving a match in highly ranked memory units, but it decreases faster with a weakly supervised assignment.
\label{fig:ProbFig}}
\vspace{-5pt}
\end{figure*}

\mypar{Better hypothesis test.}
Figure~\ref{fig:ProbFig} shows that highly ranked memory units are very likely true positives containing at least one matching vector. On the contrary, with the random assignment, a positive memory unit may have a low rank. 
This means that we now can analyse the database vectors of a shorter list of memory units to find most of the matching vectors. 

\begin{figure*}
\begin{center}
	\includegraphics[height=2.1cm,trim=0cm 0cm 0cm 0cm,clip]{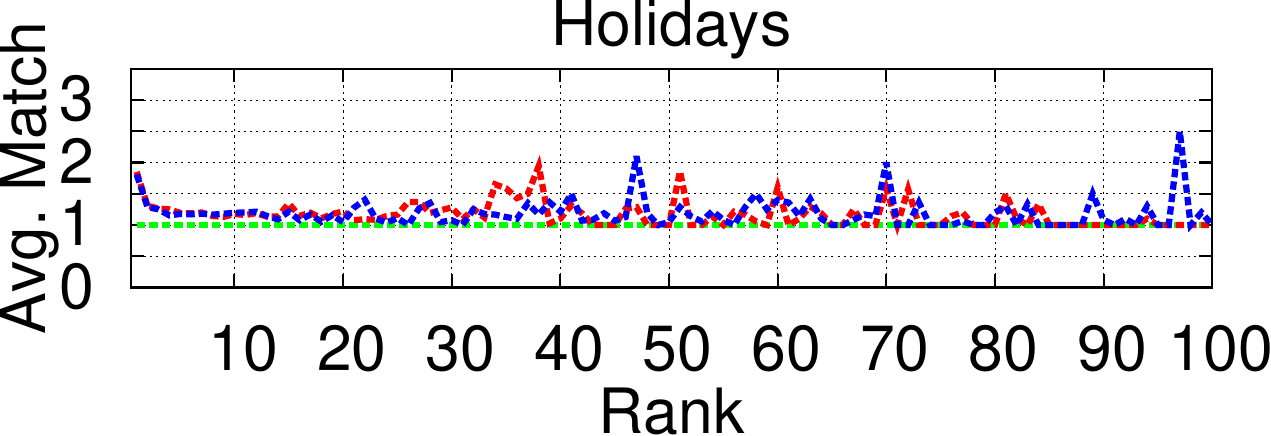}
	\hfill
	\includegraphics[height=2.1cm,trim=1cm 0cm 0cm 0cm,clip]{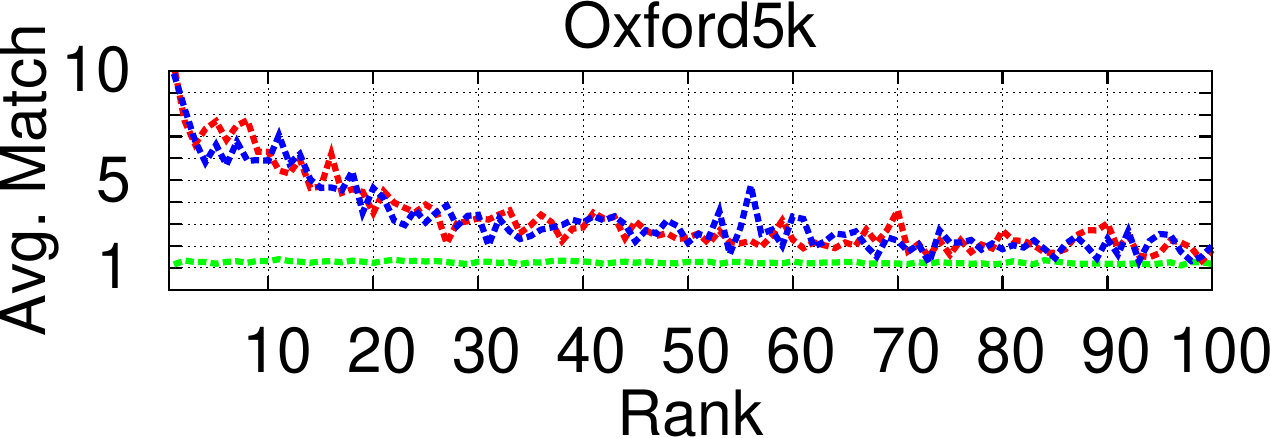}
	\hfill
	\includegraphics[height=2.1cm,trim=1cm 0cm 0cm 0cm,clip]{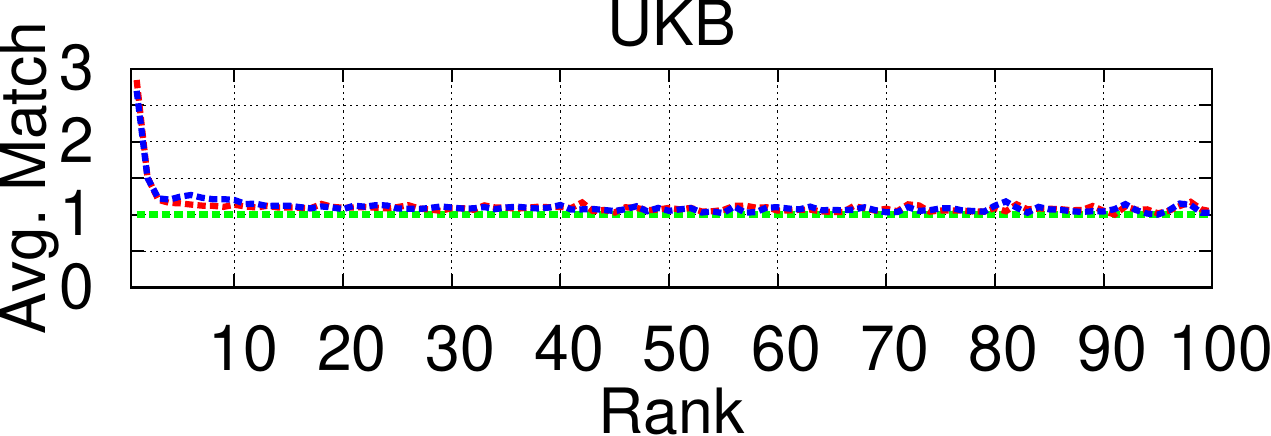}
	\hfill
\end{center}
\caption{The average number of matches given that the memory unit is positive. Green, red and blue lines correspond to random, \textit{sum} spherical k-means and \textit{pinv} spherical k-means respectively. Using random assignment, we have about 1 matching vector per memory unit. The weakly supervised assignment improves this especially for higher-ranked units.
\label{fig:MatchFig}}
\vspace{-5pt}
\end{figure*}

\mypar{More than one match.}
Another byproduct of weakly supervised assignment is that positive memory units are likely to contain more than one match, since matching vectors usually have high cosine similarity with each other. This helps the search efficiency by returning most of the matching vectors by only scanning a few positive memory units. This is also experimentally shown in Figure~\ref{fig:MatchFig}.
With the random assignment, we have almost surely at most one matching vector in each positive memory unit.

\begin{figure}
        {\includegraphics[width=0.48\textwidth]{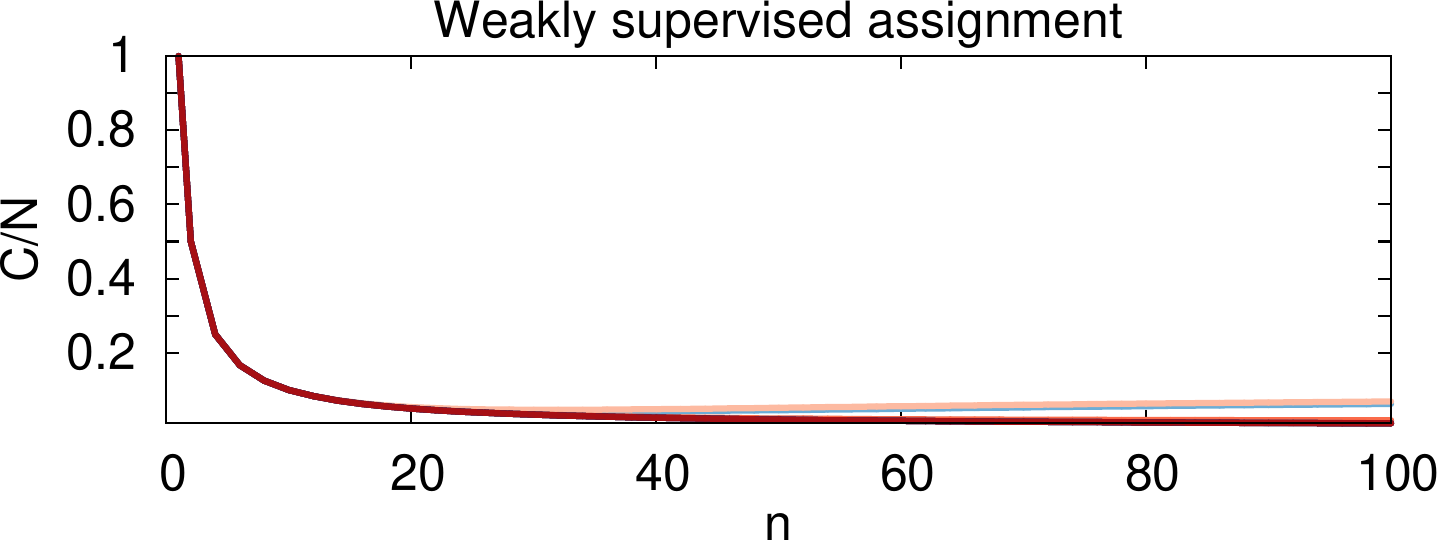}}
\caption{Ratio of the global cost by the cost of the exhaustive search $C_{\mathcal{H}_{0}}/N$ as a function of $n$ using weakly supervised assignment and synthetic data. Setup: $d=1,000$, $\epsilon = 10^{-2}$. The different curves correspond to values of $\alpha_0 \in\{0.5, 0.6 , 0.7,0.8, 0.9\}$. Red and blue lines correspond to \textit{sum} and \textit{pinv} constructions respectively. Darker shades correspond to higher $\alpha_0$. 
\label{fig:diffMKmeans}}
\end{figure}

\mypar{Imbalance factor.}
We now analyze the cost of search with weakly supervised assignment. In Eq.~\eqref{eq:cost}, we assume that each unit contains $n$ vectors. Up to this approximation, Fig.~\ref{fig:diffMKmeans} shows that both constructions $\emph{sum}$ and $\emph{pinv}$ perform better as the inner correlation increases, but more surprisingly, they perform equivalently. It is also shown that it is possible to pack many vectors into the same memory unit with weakly supervised assignment, and still obtain a low search cost.
 
In practical applications, assuming that each unit contains a constant number of vectors is no longer true with weakly supervised assignment.
This makes the analysis of the complexity more involved than~\eqref{eq:cost}.
Moreover, this is potentially problematic in some applications: the complexity and thus the runtime can change dramatically from one query to another. 

Imbalance factor is a metric to measure the impact of unbalanced clusters~\cite{JDS10a}.
It is defined as
\begin{equation}
\label{eq:imbFact}
\delta= M \sum^{M}_{i=1} {p_i}^2, 
\end{equation}
where $M$ is the number of clusters, and $p_i$ is the empirical probability that a database vector belongs to the $i$-th cluster. This is measured as frequency $p_i = n_i/N$, where $n_i$ is the cardinality of the $i$-th cluster.
Simple derivations give the following expectation and variance: $\E(n_i)=N/M$ and $\V(n_i) =(\delta-1)N^{2}/M^{2}$.
This shows that higher imbalance factor corresponds to clusters with varying sizes. This gives birth to a wide
variability of the complexity from one query to another.

\begin{figure*}[ht]
        \includegraphics[height=0.27\textwidth]{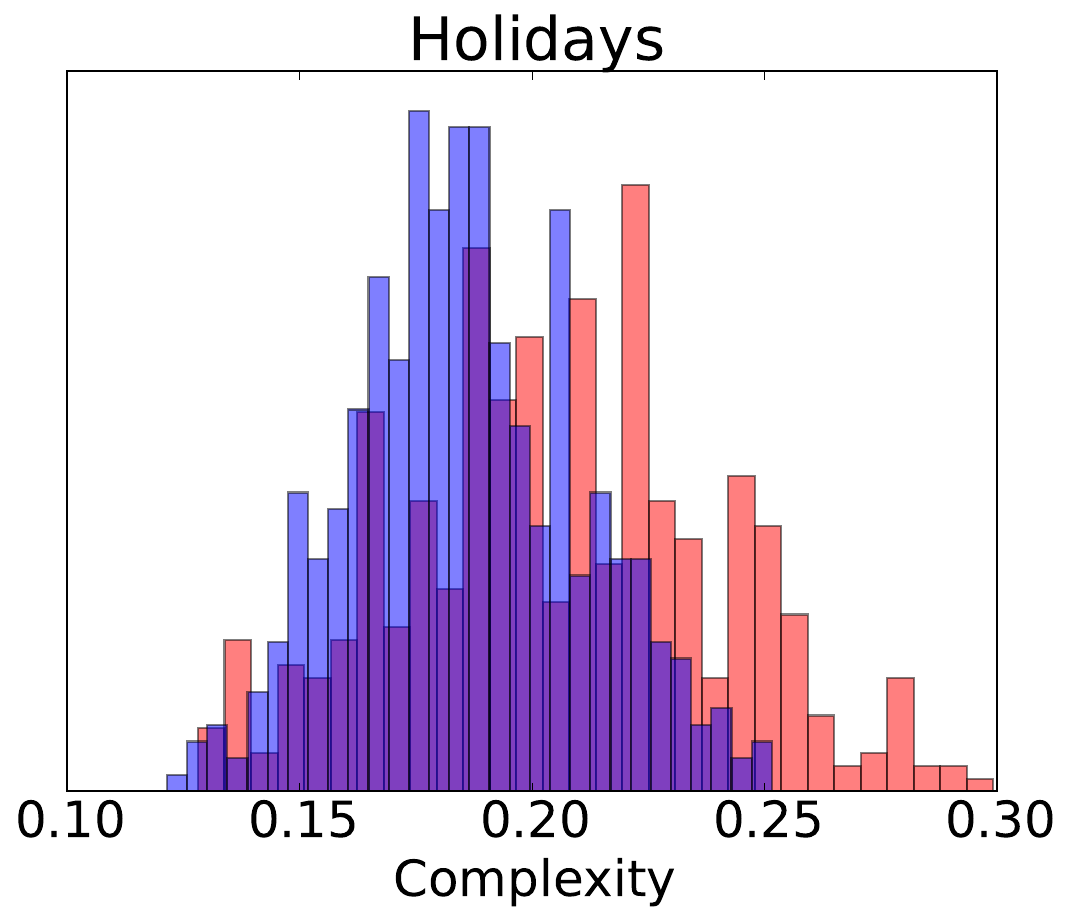}
    \hfill            
        \includegraphics[height=0.27\textwidth]{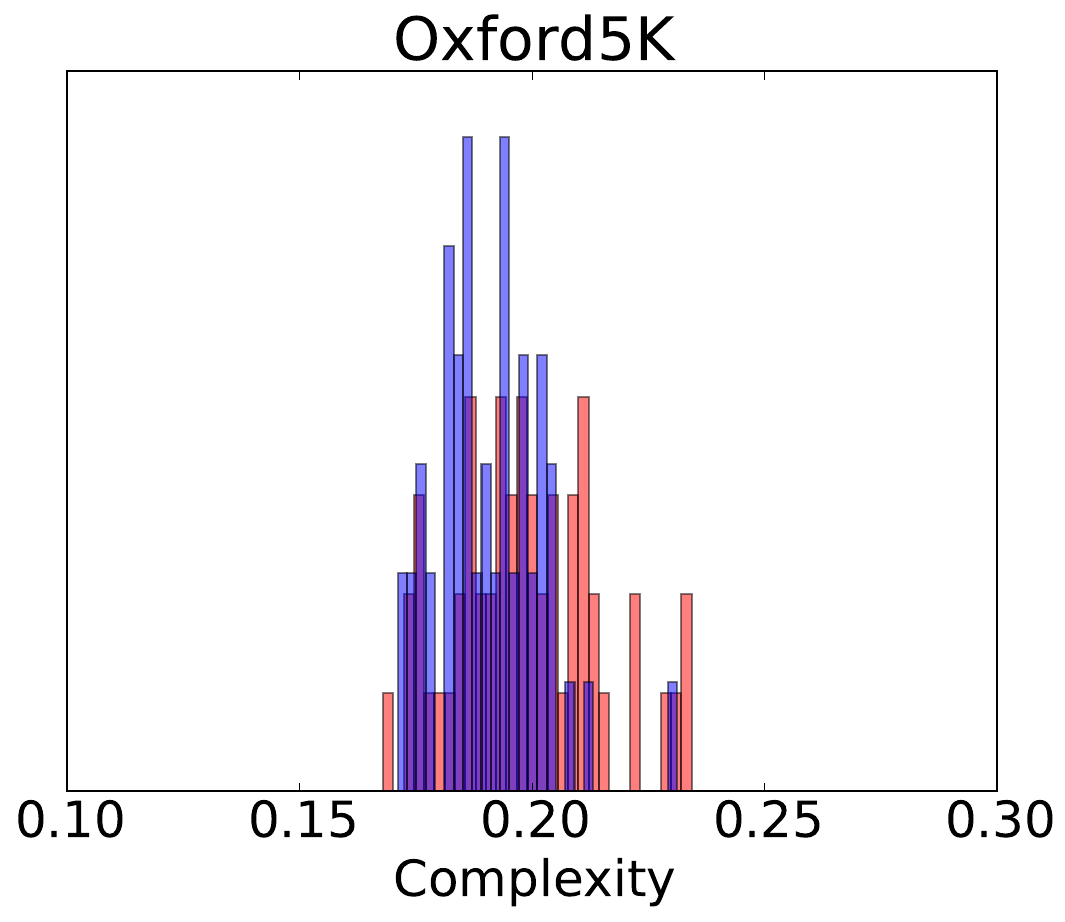}
    \hfill   
        \includegraphics[height=0.27\textwidth]{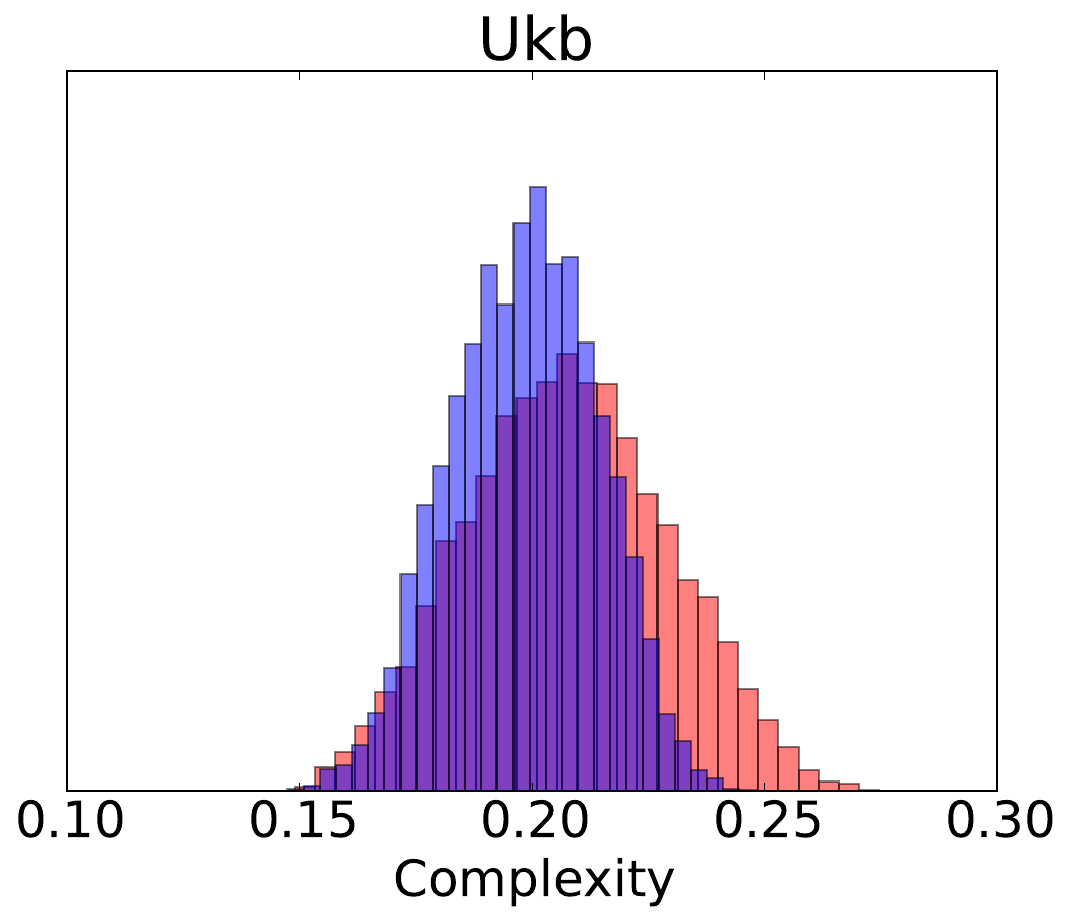}
\caption{Complexity per query for \textit{pinv} k-means (blue) and \textit{sum} k-means (red). Although the averaged complexity ratios over all queries are similar, \textit{pinv} has less variance thanks to lower imbalance factor. 
\label{fig:queryChecks}}
\end{figure*}

Table~\ref{tab:imbFactor} shows the imbalance factor for different weakly supervised assignments. It is shown that \textit{pinv} variants have more balanced cells compared to traditional \textit{sum}, making the search process more effective. The negative effect of high imbalance factor in practice is better observed in Figure~\ref{fig:queryChecks}. In this figure, the algorithm visits a fixed number of positive memory units: $7$ (Holidays), $30$ (Oxford5k), or $60$ (UKB).
This roughly gives us a complexity ratio of $C_{\mathcal{H}_{0}}(\tau) \approx 0.2$ on average.
We then show the complexity ratio per query in a histogram. It is clearly seen that the distribution for \textit{pinv} has smaller standard deviation compared to \textit{sum}, even though their means are almost the same. This makes \textit{pinv} variant of spherical $k$-means a better alternative for weakly supervised assignment.

\begin{table}
\centering{
\small
    \begin{tabular}{| l | c | r | r | r |}
    \hline
     			& sum 				& sum + norm. 	& pinv				& pinv + norm. 		\\ \hline
     Holidays	& 2.08\				& 2.17\ 			& \textbf{1.90}\				& 2.03\		\\  
     Oxford5k	& 2.76\ 				& 2.69\ 			& 2.27\				& \textbf{2.23}\		\\      
     UKB			& 2.58\ 				& 2.56\ 			& \textbf{2.06}\		& 2.09\				\\ \hline     
\end{tabular}}
\caption{Imbalance factor for different datasets using \textit{sum}, \textit{pinv} and their normalized variants of k-means. Each dataset is clustered into $M=N/10$.
\label{tab:imbFactor}}
\end{table}

%% file: applications.tex

\section{Application to image search}
\label{sec:app}

This section shows that memory vectors perform extremely well on typical computer vision benchmarks.
We assume two scenarios: closed-datasets in Section~\ref{sec:closed}, and large-scale and streaming data in Section~\ref{sec:stream}.

Whereas the datasets were already introduced in Sect.~\ref{sub:ExpSetup}, let us describe the measure of performances.
We follow the standard image retrieval protocol where each image is represented by a feature vector and the ground truth is now based on the visual similarity. The goal is to return visually similar images for a given query image. The similarity of two images is measured by the cosine of their descriptor vectors, and the images are ordered accordingly. We adopt the performance measure defined for each benchmark: mAP (mean average precision) which measures the area under the precision-recall curve~\cite{PCISZ07} or 4-recall@4, which is the average number of correct images in the top-4 positions. 

As for the complexity, we first measure the similarities between the query and $M$ memory vectors.
We compare these similarities with a given threshold $\tau$. Then, we re-rank all the vectors in positive memory units according to their similarities with the query vector. 
We characterize the complexity of the search per database vector by:
\begin{equation}
C_{\mathcal{H}_{1}}(\tau) = M + \sum_{i: \y^{\top}\m_{i}>\tau} n_{i},
\end{equation}
where $n_{i}$ is the number of database vectors in the $i$-th memory unit.
We measure the complexity ratio $C_{\mathcal{H}_{1}}(\tau)/N$ and the retrieval performance for different values of the threshold $\tau$. For large $\tau$, no memory unit is positive, resulting in $C_{\mathcal{H}_{1}}(\tau)/N = M $ and no candidate is returned. As $\tau$ decreases, more memory units trigger reranking. 

\subsection{Closed dataset}
\label{sec:closed}
\begin{figure*}
        \includegraphics[height=0.23\textwidth]{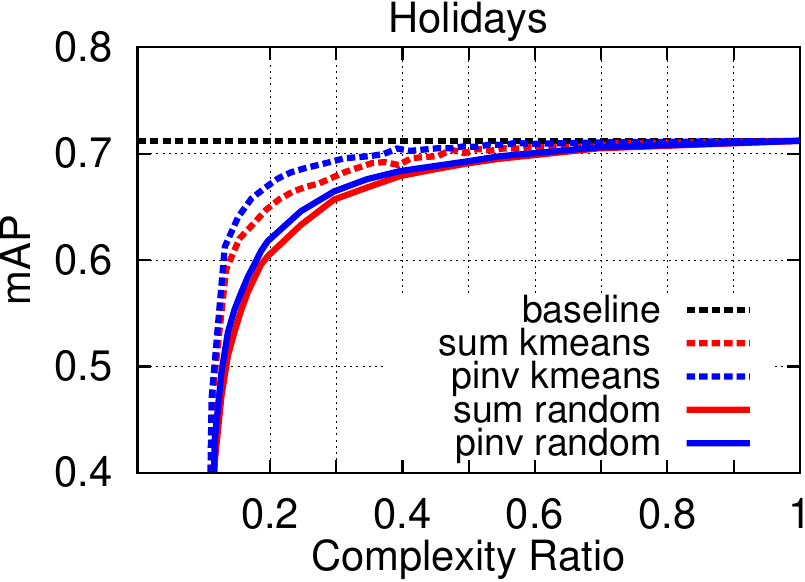}
    \hfill            
        \includegraphics[height=0.23\textwidth]{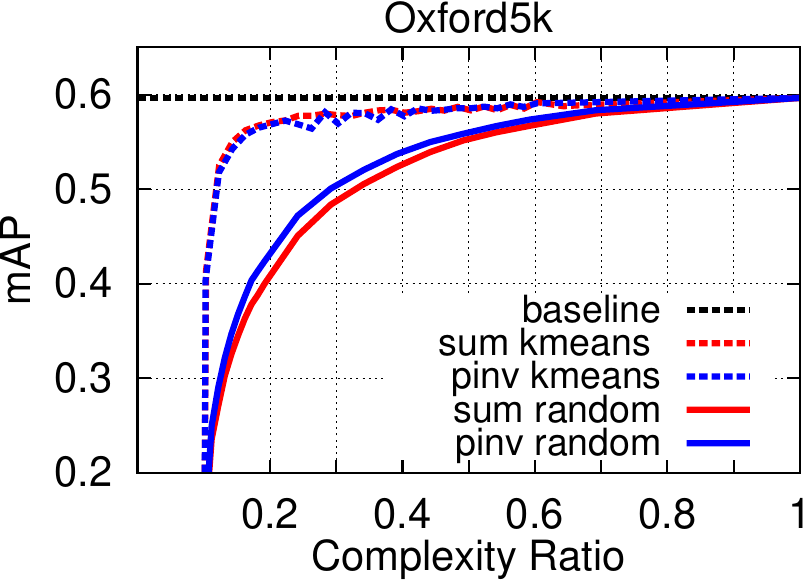}
    \hfill   
        \includegraphics[height=0.23\textwidth]{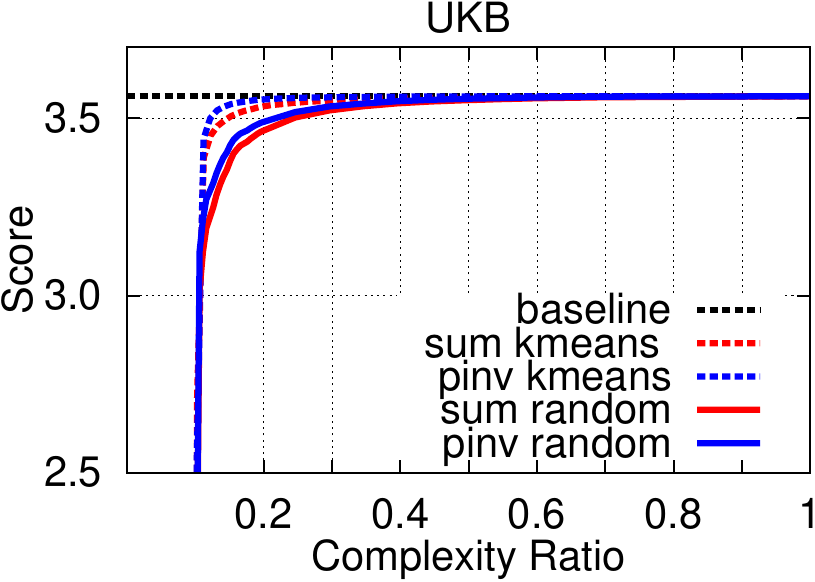}
\caption{Image retrieval performance using visual similarity ground truth. K-means variants bring significant improvement.
\label{fig:retInit}}
\end{figure*}

\begin{figure*}
        {\includegraphics[width=0.48\linewidth]{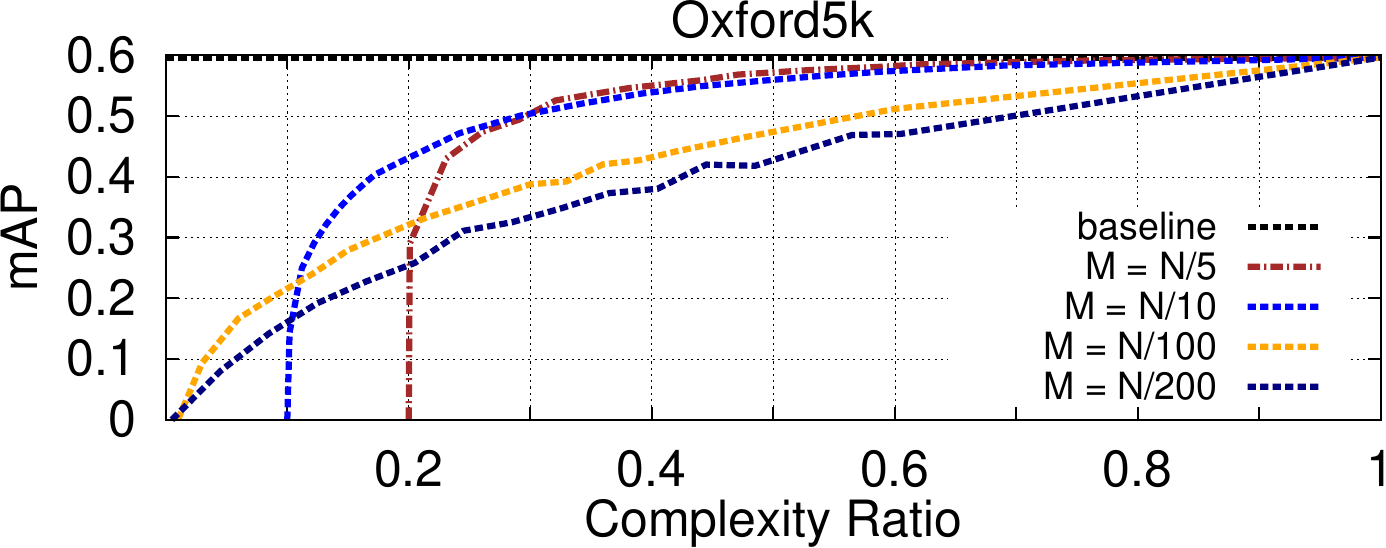}}
    \hfill            
       \includegraphics[width=0.48\linewidth]{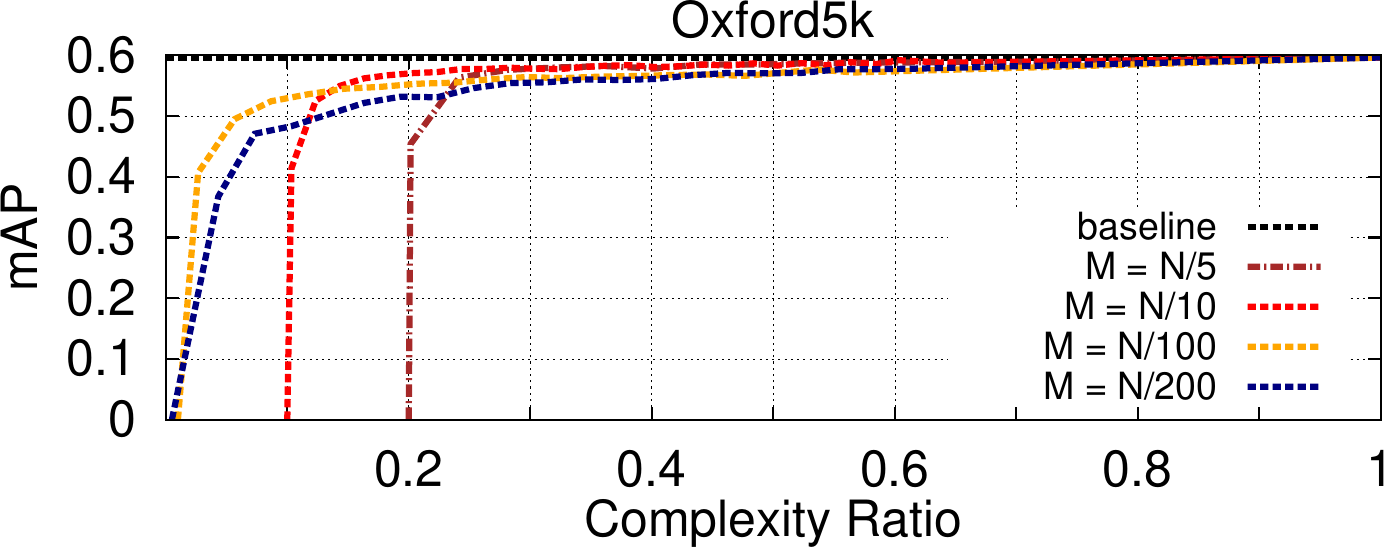}
\caption{Search performance using random assignment (left) and weakly supervised assignment (right). This last option uses fewer memory units and still obtains a good search performance, which is not possible with random assignment.
\label{fig:diffMOxford}}
\end{figure*}

Recall from Section~\ref{sec:coma} that weakly supervised assignment provides better approximate search than a random assignment. This is confirmed for the image search benchmark in Figure~\ref{fig:retInit}. Additionally, we show that it is possible to pack more vectors in a memory unit using weakly supervised assignment in Figure~\ref{fig:diffMOxford}. We use this approach (spherical $k$-means with \textit{pinv}) for the rest of our experiments.

\begin{figure}
\centering
 \includegraphics[width=0.95\linewidth,trim=0cm 1.5cm 0cm 0cm,clip]{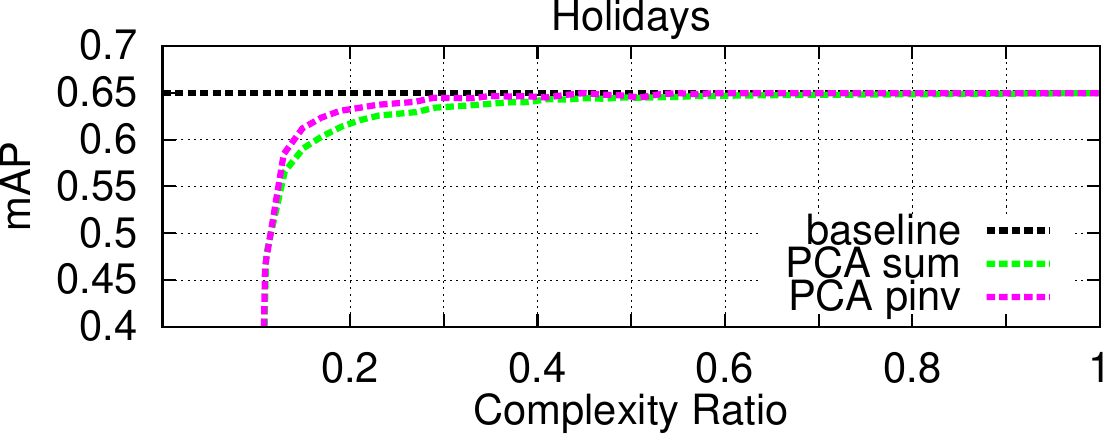} \\
 \includegraphics[width=0.95\linewidth,trim=0cm 1.5cm 0cm 0cm,clip]{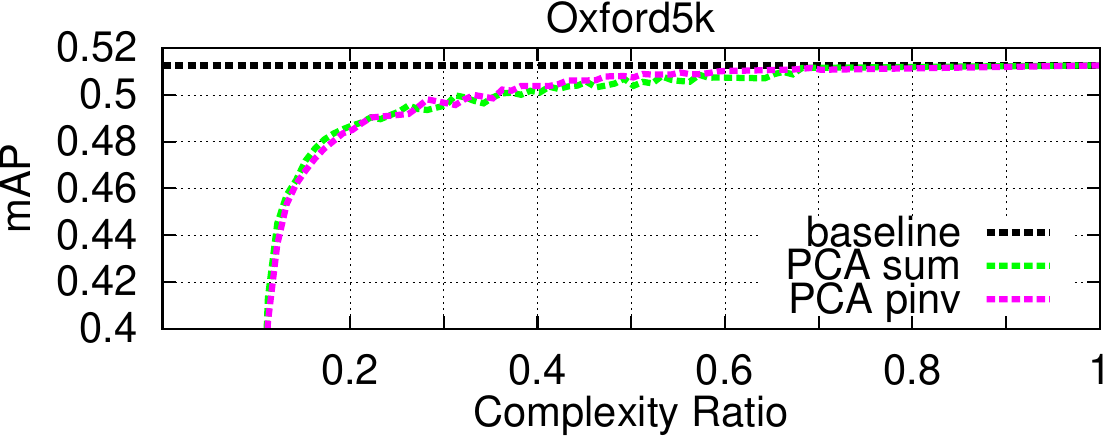} \\
\includegraphics[width=0.95\linewidth,trim=0cm 0cm 0cm 0cm,clip]{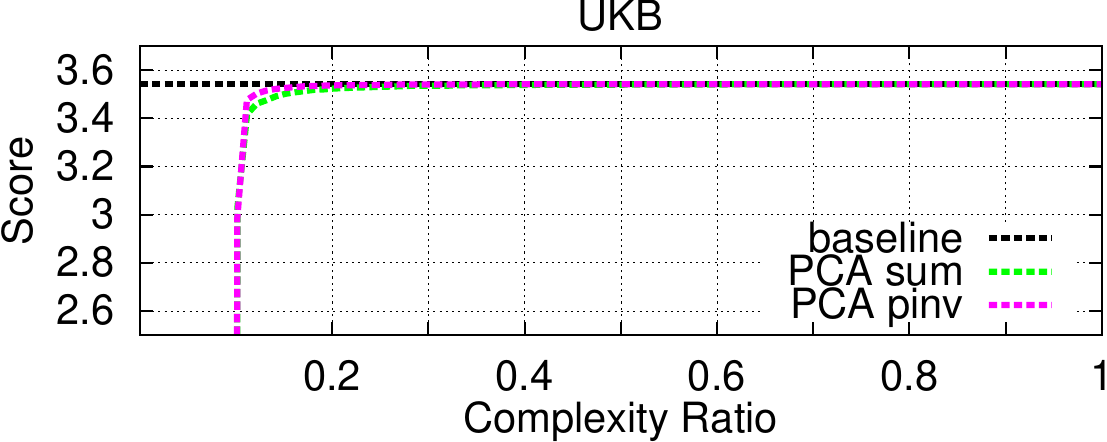}
\caption{The performance of memory vectors after PCA dimensionality reduction with $d'=1024$.
\label{fig:retDim}}
\end{figure}

\begin{figure}
\centering
 \includegraphics[width=0.95\linewidth,trim=0cm 1.5cm 0cm 0cm,clip]{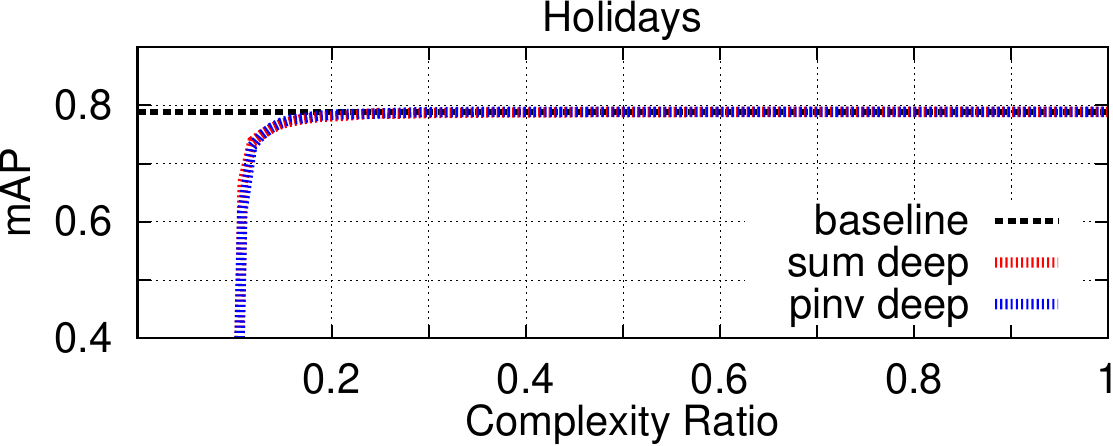} \\
 \includegraphics[width=0.95\linewidth,trim=0cm 1.5cm 0cm 0cm,clip]{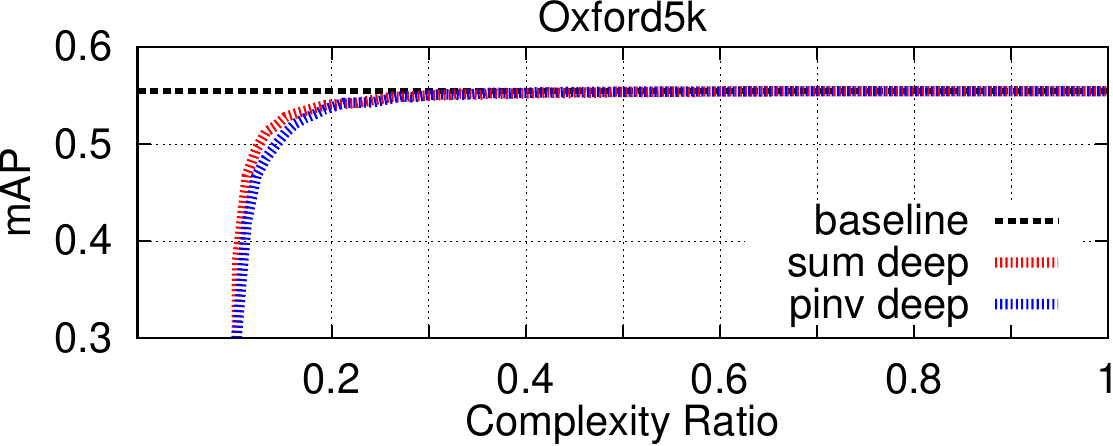} \\
\includegraphics[width=0.95\linewidth,trim=0cm 0cm 0cm 0cm,clip]{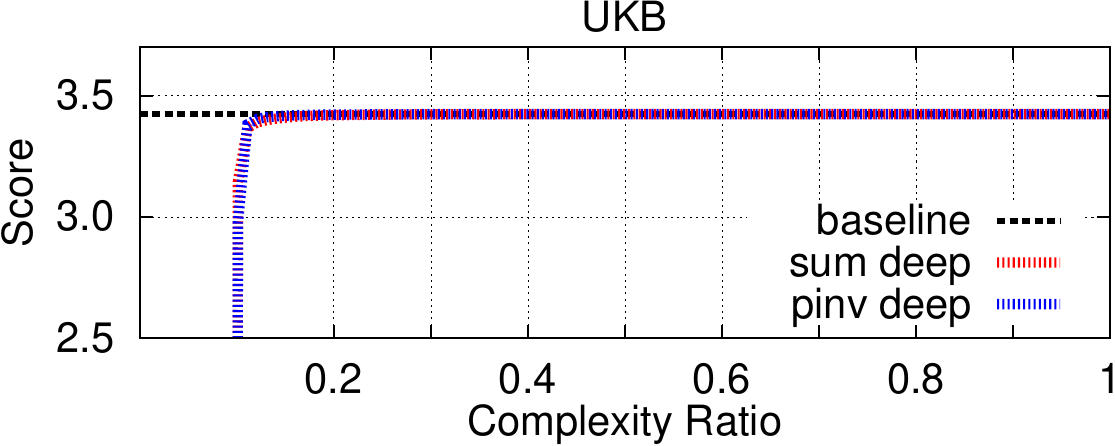}
\caption{Image retrieval performance with deep learning features ($d=4096$), as trained by Babenko \etal~\cite{BSCL14}. Features for Holidays and Oxford5K are retrained on the Landmarks dataset, whereas the ones for UKB are trained on ILSVRC. 
\label{fig:retDeep}}
\end{figure}

\mypar{The dimensionality} of the descriptor linearly impacts the efficiency of any system.
Dimensionality reduction with PCA is one way to improve this point. Our method is compatible with dimensionality reduction as shown in Fig.~\ref{fig:retDim}, where we reduce the vectors to $d'=1024$ components.
The search performance is comparable to the baseline with less computational complexity.
We also apply our method to features learned with deep learning ($d=4096$).
Fig.~\ref{fig:retDeep} shows that the reduction in complexity also applies when using high performance deep learning features.

\mypar{Compact codes }are another way to increase efficiency. We reduce the dimensionality of the vectors to $d'$=$1024$ and binarize them by taking the \textit{sign} of each component, in the spirit of cosine sketches~\cite{C02}. In the asymmetric case~\cite{GP11,JJG11}, only memory vectors and dataset vectors are binarized, whereas in the symmetric case query vectors are also binarized during the query time.

\begin{figure}[ht]
\centering
 \includegraphics[width=0.95\linewidth,trim=0cm 1.5cm 0cm 0cm,clip]{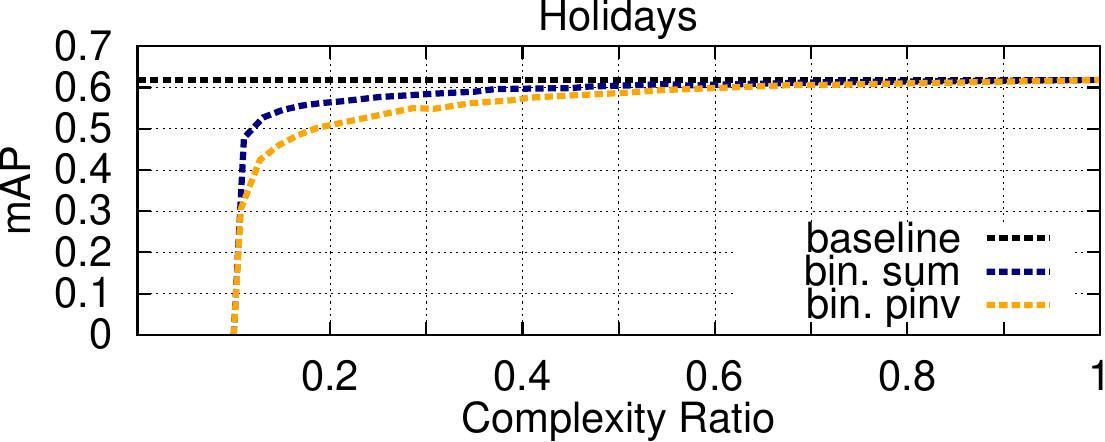} \\
 \includegraphics[width=0.95\linewidth,trim=0cm 1.5cm 0cm 0cm,clip]{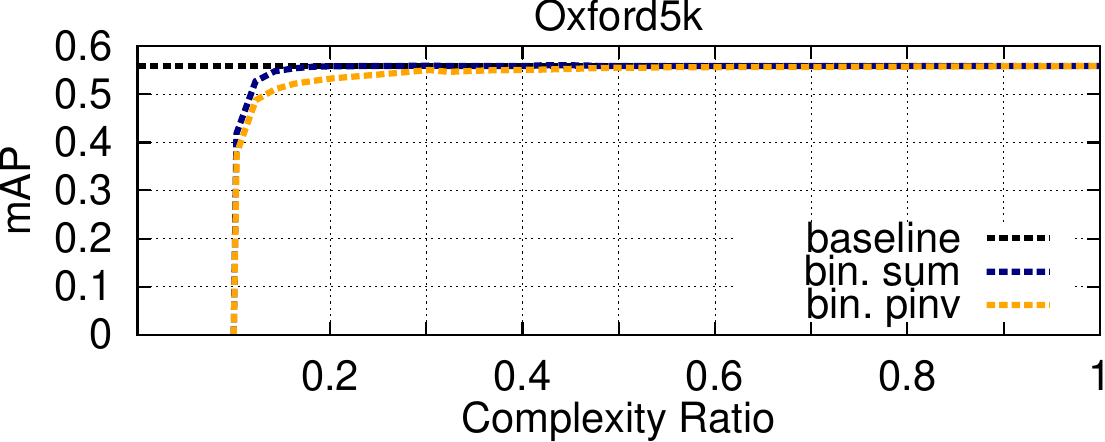} \\
\includegraphics[width=0.95\linewidth,trim=0cm 0cm 0cm 0cm,clip]{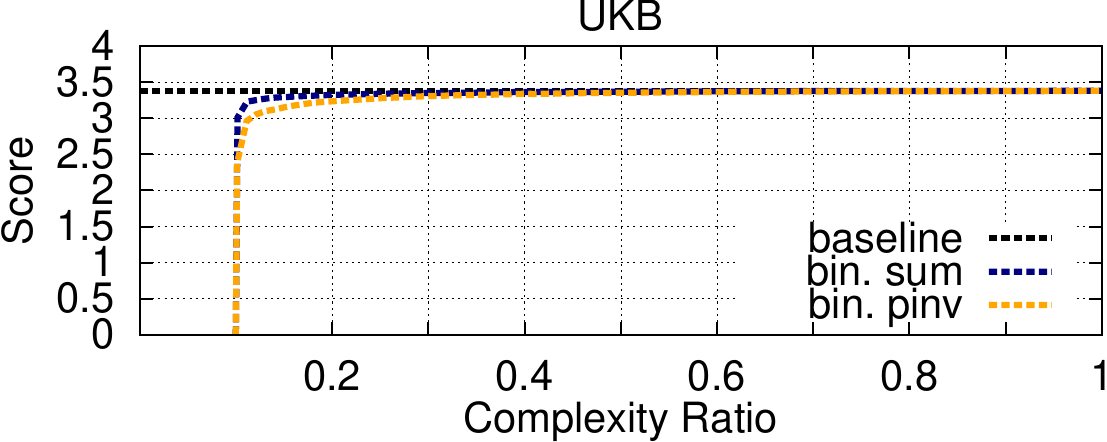}
\caption{Experiments with binary codes after PCA reduction to $d'=1024$. The quantization is symmetric: real query vectors are binarized and then compared to binary memory vectors. 
\label{fig:retBinSym}}
\end{figure}

\begin{figure}[ht]
\centering
 \includegraphics[width=0.95\linewidth,trim=0cm 1.5cm 0cm 0cm,clip]{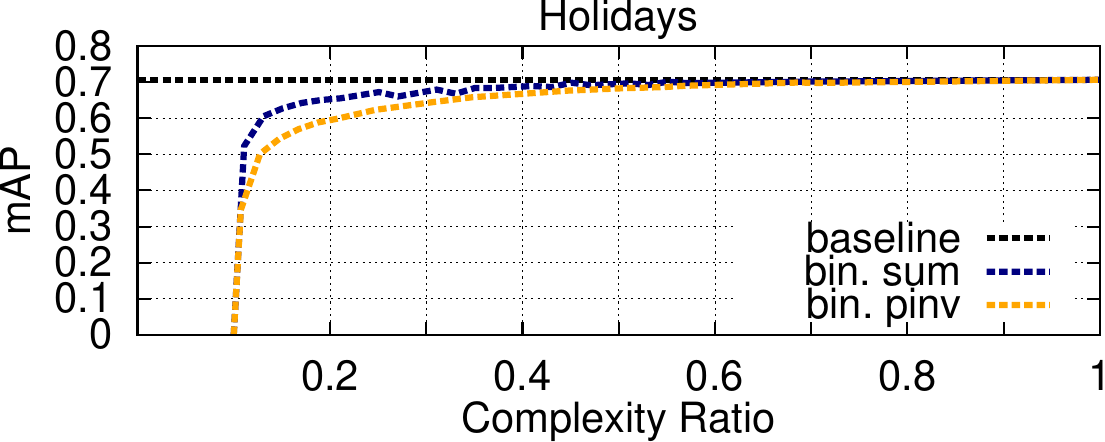} \\
 \includegraphics[width=0.95\linewidth,trim=0cm 1.5cm 0cm 0cm,clip]{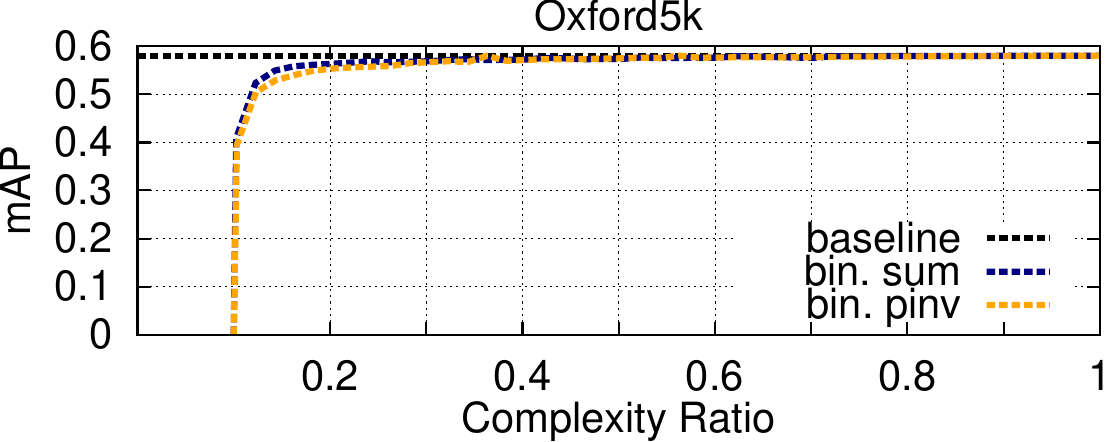} \\
\includegraphics[width=0.95\linewidth,trim=0cm 0cm 0cm 0cm,clip]{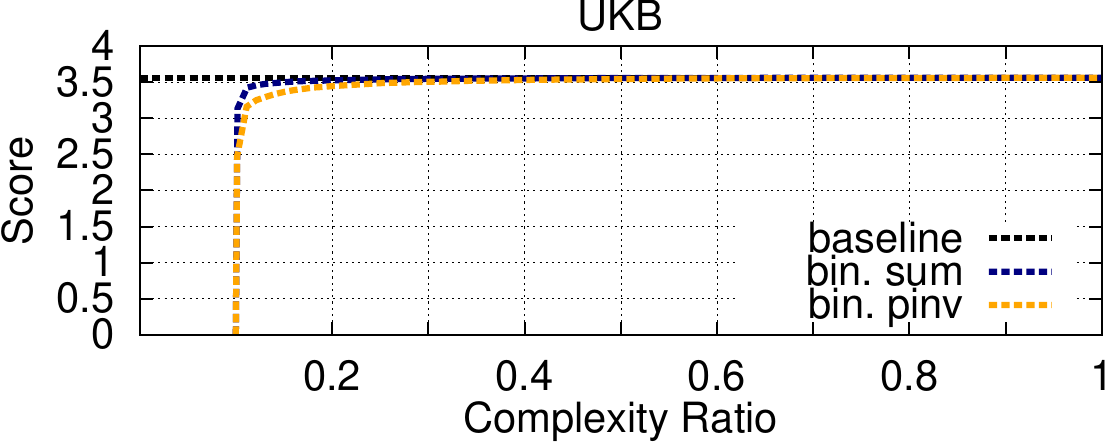}
\caption{Experiments with binary codes after PCA reduction to $d'=1024$. The quantization is asymmetric: real query vectors are compared to binary memory vectors.  
\label{fig:retBinAsym}}
\end{figure}

Figures \ref{fig:retBinSym} and \ref{fig:retBinAsym} show the performance when using compact binary codes.
For the symmetric case, the \textit{sum} method seems to perform better than \textit{pinv} on the Holidays and Oxford5k datasets. In the asymmetric case, both methods perform similarly.
In all cases, we achieve convergence to the baseline with a complexity ratio well below 1. Implementation efficiency is further improved in the symmetric case by using the Hamming distance calculation instead of dot product. 

\mypar{Comparison with FLANN}~\cite{ML14}.
Running the FLANN algorithm on the Holidays+Flickr1M dataset reveals that the convergence to the baseline is achieved with a speedup of 1.25, which translates to a complexity ratio of around 0.8.
We can achieve similar performance with a complexity ratio of only 0.3. This confirms that FLANN is not effective for high-dimensional vectors.
In this experiment, we use the autotune option of the FLANN library, and set target\_precision~= 0.95, build\_weight~= 0.01, and memory\_weight~= 0.

\mypar{Execution time.}
 We have shown that we get close to baseline performance while executing significantly fewer operations. We now measure the difference in execution time under a simple setup: $d=1024$ and $N=10^{6}$ dataset vectors. An average dot product calculation between the query and all dataset vectors is 0.2728$s$. With $N/10$ memory vectors and $\approx 100k$ vectors in positive memory vectors, the execution time decreases to 0.0544$s$. We improve the efficiency even further with symmetric compact codes and Hamming Distance computation: the execution time becomes 0.0026$s$. Our method is parallelized for further improvement.

\subsection{Large scale and streaming data}
\label{sec:stream}
We conduct large scale experiments on Holidays+Flickr1M and Yahoo100M datasets. The main advantage of our approach is its compatibility with large scale and streaming data, where pre-clustering the data may not be possible. More specifically, we assume that we have streaming images which we would like to index. As the size of the data keeps growing continuously, it is not possible to apply traditional $k$-means in such a scenario. We investigate two different approaches: random assignment and weakly supervised assignment over mini-batches.

\mypar{Online indexing } assumes that we would like to index items in streaming data as they become available. In such case, the random assignment is applicable provided that the successive vectors in the stream are independent. 

\begin{figure*}
        \includegraphics[height=0.25\textwidth]{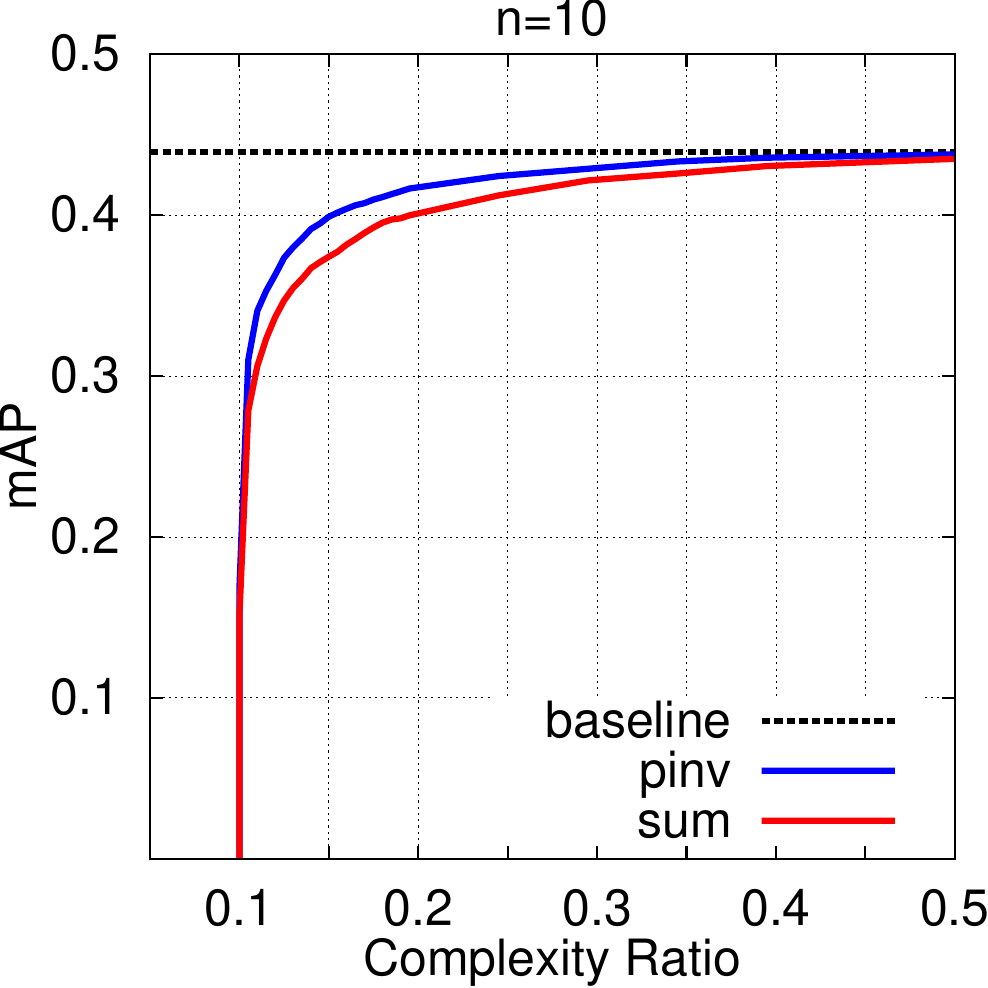}
    \hfill            
        \includegraphics[height=0.25\textwidth,trim=0.7cm 0cm 0cm 0cm,clip]{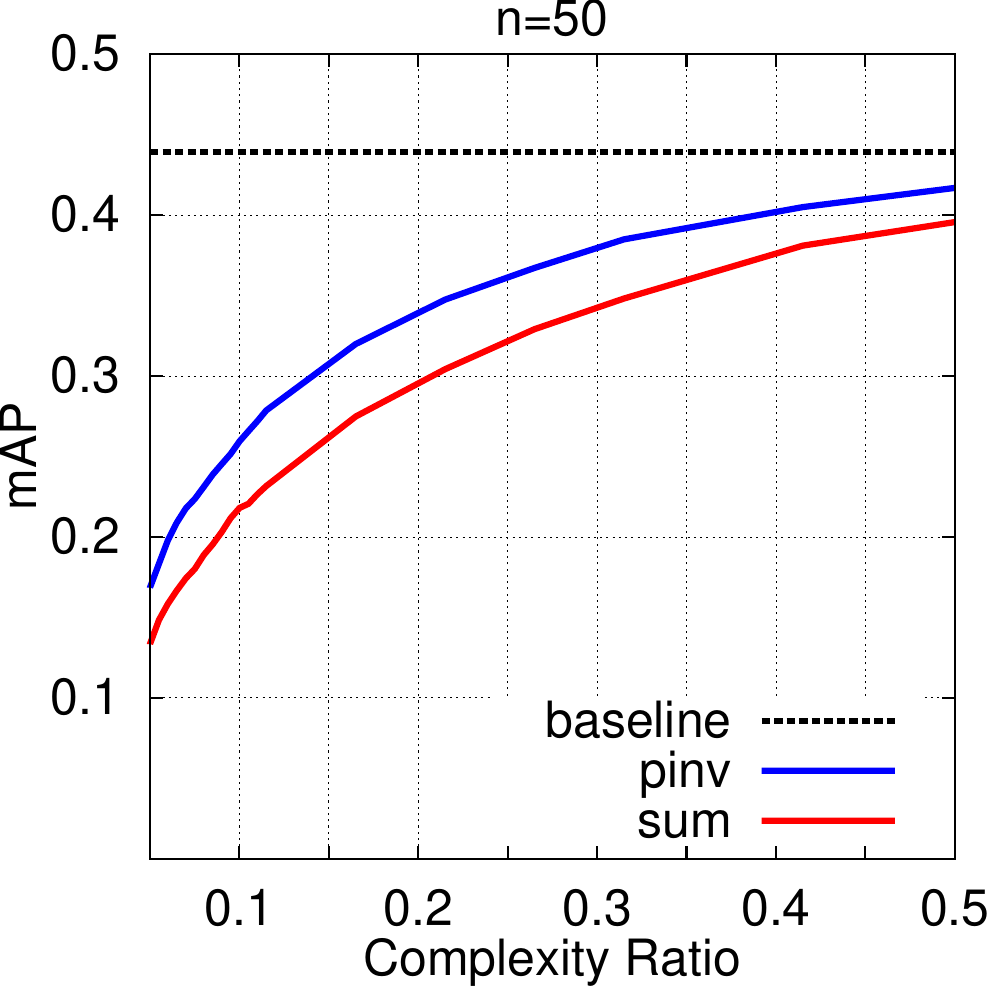}
    \hfill   
        \includegraphics[height=0.25\textwidth,trim=0.7cm 0cm 0cm 0cm,clip]{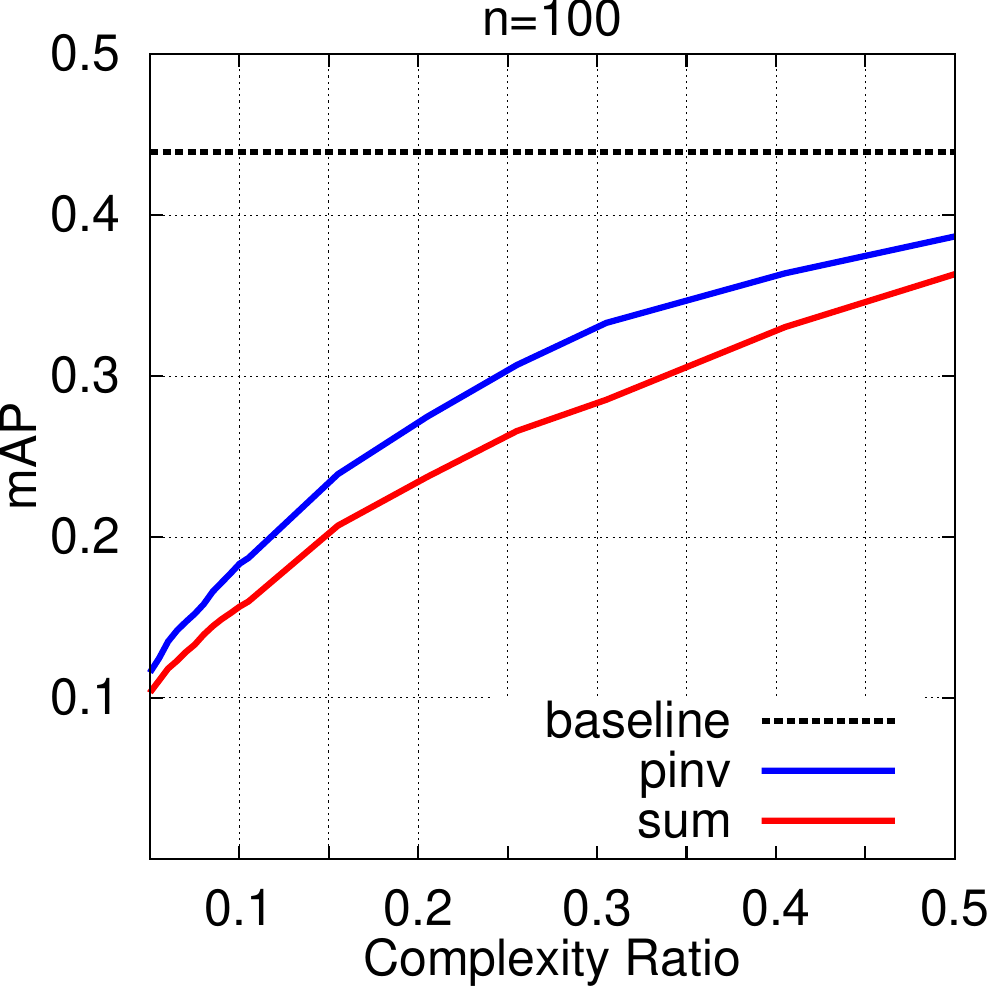}
    \hfill   
        \includegraphics[height=0.25\textwidth,trim=0.7cm 0cm 0cm 0cm,clip]{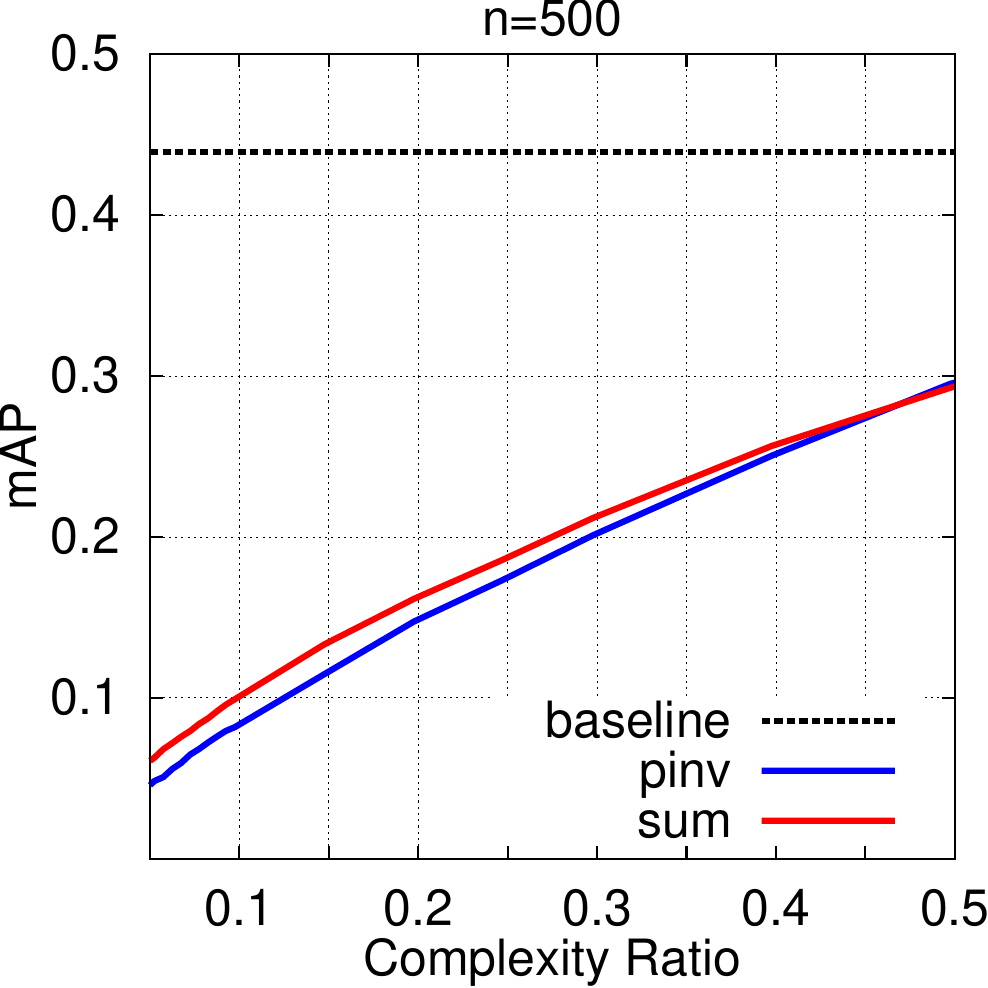}
    
\caption{Image retrieval performance in Holidays+Flickr1M with different memory unit size $n$. The data is streamed in this scenario, and the memory untis are created randomly.
\label{fig:FlickrRandom}}
\end{figure*}

Figure~\ref{fig:FlickrRandom} shows the image retrieval performance based on random assignment with different group sizes $n$. With $n=10$, the performance is close to the baseline while performing roughly three times fewer vector operations than exhaustive search. On the other hand, larger groups make it possible to have smaller complexity ratio with a degrading effect on the quality of search, since the scores obtained from memory vectors are noisier (see~\eqref{eq:sumfpfn} and~\eqref{eq:pinvfpfn}). The \textit{pinv} construction performs better than \textit{sum} in all cases except for a very large memory units of size $n=500$, where the quality of search is low in general.

\begin{figure}[t]
\centering
 \includegraphics[width=0.95\linewidth,trim=0cm 0cm 0cm 0cm,clip]{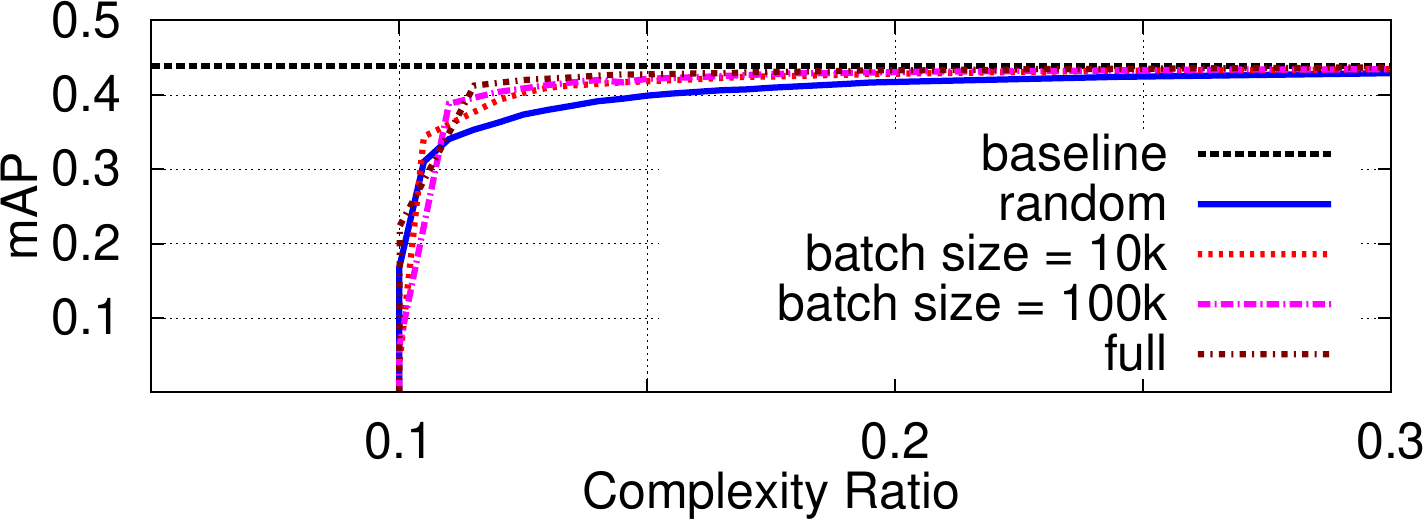}
\caption{Comparison of random and weakly supervised assignments over batches in a large-scale setup. Option \textit{full} corresponds to a unique batch of size $10^{6}$. We run each experiment multiple times.
\label{fig:retFlickr1Mbatch}}
\end{figure}

\smallskip
\mypar{Batch assignment.} The alternative approach runs weakly supervised assignment on small batches. A \emph{batch} spherical $k$-means is the same as the regular spherical $k$-means discussed in previous sections. We do not cluster the whole dataset at once because this would not be tractable with large collections. Instead, we randomly divide the dataset into batches of the same size and run a weakly supervised assignment separately for each batch. Fig.~\ref{fig:retFlickr1Mbatch} shows that this strategy improves the performance while keeping the complexity of the clustering algorithm manageable. 

\begin{figure}[t]
\centering
 \includegraphics[width=0.95\linewidth,trim=0cm 0cm 0cm 0cm,clip]{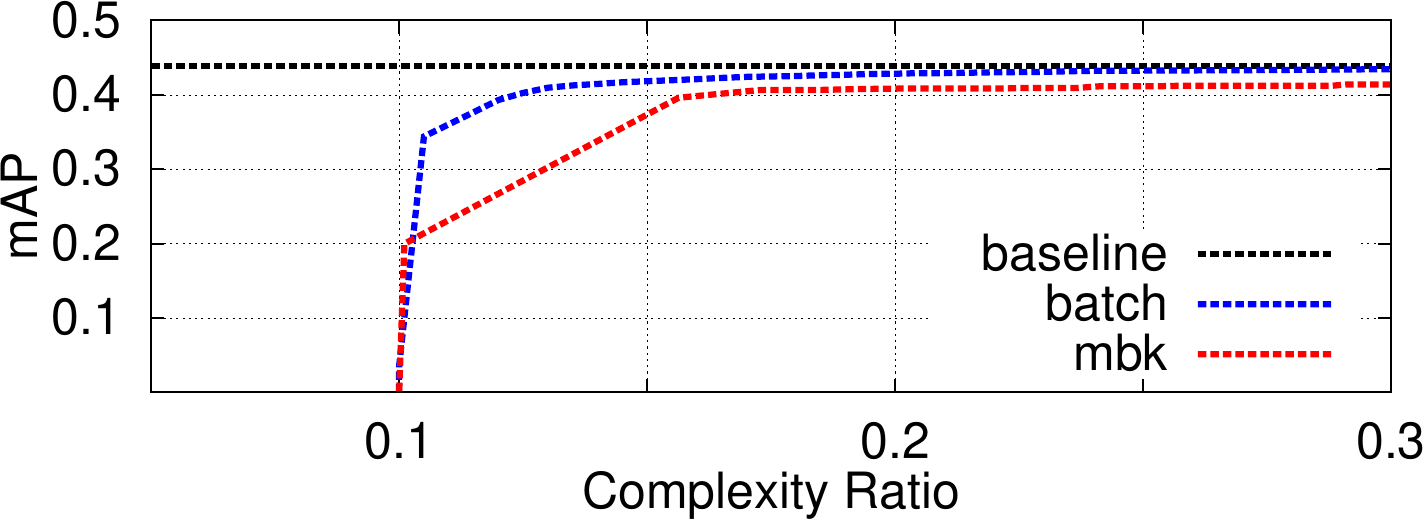}
\caption{Comparison of our \textit{batch spherical k-means} approach with the mini-batch k-means algorithm (mbk) ~\cite{S10}. The batch size equals $10$k.
\label{fig:flickrMinibatch}}
\end{figure}

We compare our approach to a well-known mini-batch $k$-means algorithm~\cite{S10} (referred to as mbk) as implemented by~\cite{PVGM+11}. We compare both strategies using a batch size of $10$k, and show the image retrieval performance for different complexity ratio in Figure~\ref{fig:flickrMinibatch}. 

The first observation is that the plot of mbk is not smooth. This is due to the clusters being unbalanced when using such mini-batch approaches. In fact, when we measure the imbalance factor~\eqref{eq:imbFact}, we obtain $\delta = 183\pm8$. As a result, few clusters contain a large number of dataset vectors, and when a cluster is accessed, the cost of the verification step becomes expensive. On the contrary, the imbalance factor observed using our batch spherical $k$-means is only $2.47\pm0.01$, resulting in a more efficient verification step for positive memory units.

\begin{figure}
\centering
        \includegraphics[height=4cm]{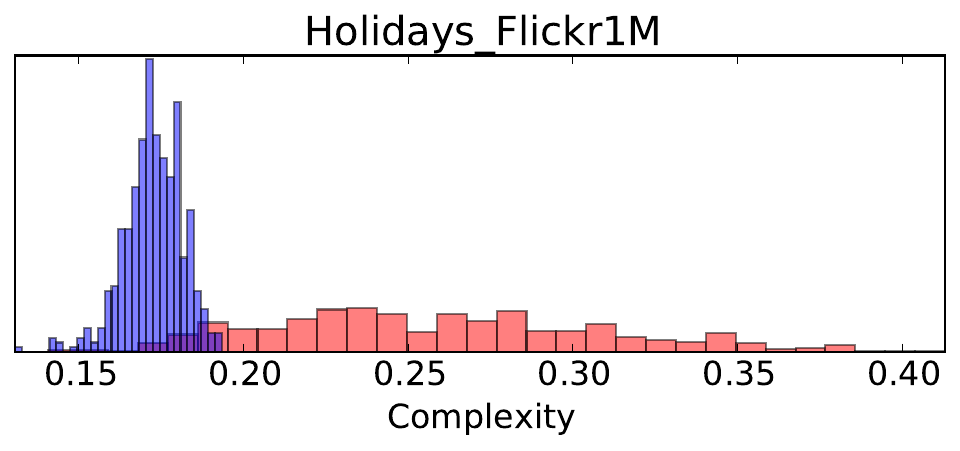}
\caption{Complexity of each query for Holidays+Flickr1M using our batch spherical k-means (blue) and mbk (red). Our scheme has less complexity on average and less variance.
\label{fig:queryChecksFlicrk1M}}
\end{figure}

The complexity ratio per query for the two methods can be seen in Figure~\ref{fig:queryChecksFlicrk1M}. When we set the number of positive memory units to $3500$, the mean complexity ratio for our batch spherical $k$-means approach is 0.17, with a standard deviation of 0.01. On the other hand, when mbk is used, the mean increases to 0.26 with a standard deviation of 0.51.

\begin{figure}
\centering
        \includegraphics[height=3cm]{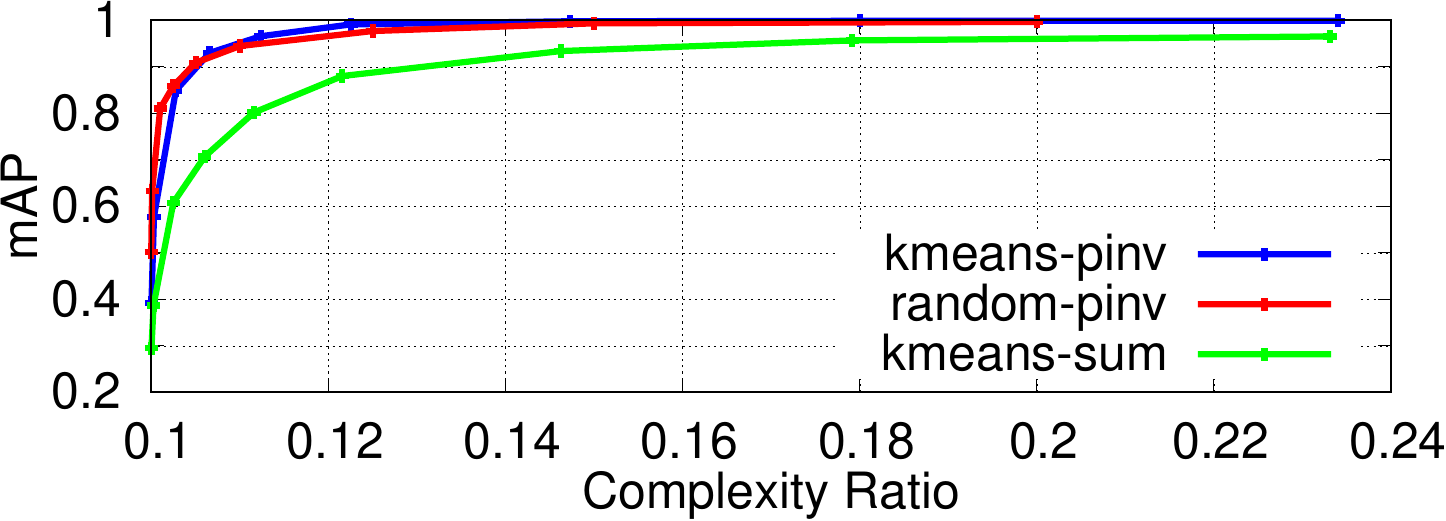}
\caption{Comparison of \textit{pinv} and \textit{sum} based methods in Yahoo100M. Our \textit{pinv}-based k-means variant achieves baseline performance with only 0.12 complexity ratio.
\label{fig:yahoo100MFig}}
\end{figure}

Finally, we apply our batch assignment strategy to the Yahoo100M, which consists of 97.6 million vectors. We divide the dataset into batches of 100k, and run three different indexing strategies for each batch: random assignment with \textit{pinv}, \textit{pinv}-based spherical k-means and \textit{sum}-based spherical k-means. This dataset does not have manually annotated ground truth or designated queries, therefore We use an existing evaluation protocol~\cite{IRF16}.  Dataset vectors are considered a match if they have a similarity of $0.5$ with the query vector. 1000 query vectors are randomly chosen and those that have 0 or more than 1000 matches are filtered out. In the end, we have 112 queries, with each query having 11.4 matches on average. 

We present our results in Figure~\ref{fig:yahoo100MFig}. Since we have $N/10$ memory vectors for each batch, a lower bound for the complexity ratio is 0.1. We see that \textit{pinv}-based methods converge to the baseline much faster than \textit{sum}. Using our \textit{pinv}-based k-means variant, we achieve the same performance as baseline with 0.12 complexity ratio. This saves us almost 90 million vector operations at query time.

%% file: appendix.tex
\def\M{\mathbf{M}}
\def\N{\mathbf{N}}
\def\tr{\mathsf{tr}}
\def\S{\mathbf{S}}
\def\D{\mathbf{D}}
\def\s{\mathbf{s}}

\section{Analysis}
\label{sec:analysis}
In this section, we provide the theoretical analysis of the methods presented in this paper.

\subsection{Distribution of a scalar product}

\label{sec:pdf}
Let $\Y$ be a random vector uniformly distributed on the unit hypersphere in $\real^{d}$ ($\|\Y\|=1$), and $\m$ a fixed vector. This section studies the distribution of $S=\ip{\Y}{\m}$. To generate $\Y$, we can first generate a multivariate Gaussian vector $\mathbf{G}=(G_{1},\ldots,G_{d})^{\top}\sim\mathcal{N}(\mathbf{0},\mathbf{I}_{d})$, where $\mathbf{I}_{d}$ is $d \times d$ identity matrix, and set $\Y = \mathbf{G}/\|\mathbf{G}\|$. This means that $G_{i}$ are i.i.d. Gaussian distributed: $G_{i}\sim\mathcal{N}(0,1)$. Without loss of generality, using symmetry of the Euclidean norm, assume that $\m=(\|\m\|,0,\ldots,0)$. This simplifies into
\begin{equation}
S=\ip{\Y}{\m}=\|\m\|\frac{G_{1}}{\sqrt{\sum_{i=1}^{d}G_{i}^{2}}}.
\end{equation}
Obviously, $-\|\m\|\leq S\leq \|\m\|$ so that its cdf $F_{S}(s)$ equals 0 if $s\leq -\|\m\|$, and 1 if $s\geq\|\m\|$.

For all $s\in[0,\|\m\|]$, $\P(S\geq s)$ is the probability that vector $G$ belongs to the convex cone pointing in the direction $\m$ and with angle $\cos^{-1}(s/\|\m\|)$
By a symmetry argument (replacing $\m$ by $-\m$), $\P(S\leq-s)=\P(S\geq s)$, giving 
\begin{align}
F_{S}(-s) = 1 - F_{S}(s),\,\forall 0\leq s\leq \|\m\|.
\end{align}
It also implies that 
\begin{equation}
\P(S^{2}\geq s^{2})=\P(S\geq s) + \P(S\leq -s)=2(1 - F_{S}(s)).
\end{equation}
Going back to $\mathbf{G}$, we can write
\begin{equation}
\begin{aligned}
\P(S^{2}\geq s^{2})&=\P\left(\frac{G_{1}^{2}}{\sum_{i=1}^{d}G_{i}^{2}}\geq \tau^{2}\right)\\
                   &=\P\left(\frac{G_{1}^{2}}{\sum_{i=2}^{d}G_{i}^{2}}\geq \frac{\tau^{2}}{1-\tau^{2}}\right),
\end{aligned}
\end{equation}
with $\tau = s/\|\m\|$. By definition, the random variable $U=(d-1)G_{1}^{2}/\sum_{i=2}^{d}G_{i}^{2}$ has an F-distribution $F(1,d-1)$. It follows that
\begin{align}
\P(S^{2}\geq \tau^{2}\|\m\|^{2}) = 1-I_{\tau^{2}}(1/2,(d-1)/2),
\end{align}
where $I_{x}(a,b)$ is the regularized incomplete beta function. 
In the end, $\forall s\in[0,\|\m\|]$, we have
\begin{align}
\label{eq:cdf_score}
F_{S}(s)=\left(1+I_{\frac{s^{2}}{\|\m\|^{2}}}\left(\frac{1}{2},\frac{d-1}{2}\right)\right)/2.
\end{align}

By taking the derivative of this expression and accounting for symmetry, the pdf is found to be:
\begin{align*}
f_{S}(s)=\frac{(1-s^{2}/\|\m\|^{2})^{\frac{d-3}{2}}}{\|\m\|B(1/2,(d-1)/2)}, \forall s, -\|\m\|\leq s\leq\|\m\|
\end{align*}
where $B(a,b)$ is the beta function. This implies that $\E[S]=0$ and
\begin{equation*}
\begin{aligned}
\V[S] &= \frac{\|\m\|^{2}}{B\left(\frac{1}{2},\frac{d-1}{2}\right)}\int_{0} ^{1}a^{2}(1-a^{2})^{\frac{d-3}{2}}da \\
&= \|\m\|^{2}\frac{B\left(\frac{1}{2},\frac{d-3}{2}\right)}{B\left(\frac{1}{2},\frac{d-1}{2}\right)} =\frac{\|\m\|^{2}}{d}.
\end{aligned}
\end{equation*}


When $d\rightarrow\infty$, using the expansion~\cite[Eq.~(26)]{DiDonato:1966rm} of the regularized incomplete beta function in~\eqref{eq:cdf_score} yields
\begin{equation}
F_{S}(s)\approx\Phi\left(\sqrt{\frac{d-1}{\|\m\|^{2}}}\frac{2s}{1+\sqrt{1-s^{2}/\|\m\|^{2}}}\right),
\end{equation}
which is approximately $\Phi(\sqrt{d/\|\m\|^{2}}s)$ for small $s$, i.e., the cdf of a centered Gaussian r.v.~with variance $\|\m\|^{2}/d$.

\subsection{Expected value of the memory unit}
\label{sec:AppExp}
We assume that the vectors $\x_{i}$ are not related to each other so that, for $n<d$, $\X$ has $n$ linearly independent columns. Then $\X^{+}=(\X^{\top}\X)^{-1}\X^{\top}$ and $\|\m^{\star}\|^{2}=\mathbf{1}_{n}^{\top}(\X^{\top}\X)^{-1}\mathbf{1}_{n}$. The $n\times n$ Gram matrix $\X^{\top}\X$ is real, symmetric, and positive semi-definite, and so it has an eigendecomposition $\U\Lambda\U^{\top}$, where $\Lambda$ is a diagonal matrix with non-negative coefficients $\{\lambda_{i}\}_{i=1}^{n}$ and $\U\U^{\top}=\U^{\top}\U=\mathbf{I}_{n}$. Moreover, $\tr(\X^{\top}\X) = \sum_{i=1}^{n}\lambda_{i}=n$ because $\x_{i}^{\top}\x_{i}=1$, for all $i$. 

The first construction $\m=\X\1_{n}$ has a square norm $\|\m\|^{2}=\1_{n}^{\top}\X^{\top}\X\1_{n}=\1_{n}^{\top}\U\Lambda\U^{\top}\1_{n}$. The second construction $\m^{\star}=\X(\X^{\top}\X)^{-1}\1_{n}$ has a norm $\|\m^{\star}\|^{2} = \1_{n}^{\top}(\X^{\top}\X)^{-1}\1_{n}=\1_{n}^{\top}\U\Lambda^{-1}\U^{\top}\1_{n}$.
It is difficult to say anything more for a given $\X$. However, if we consider $\X^{\top}\X$ as a random matrix, then
$\E[(\X^{\top}\X)^{-1}]-(\E[\X^{\top}\X])^{-1} = \E[(\X^{\top}\X)^{-1}]-\I_{n}$ is positive semi-definite~\cite{Groves:1969qy}. This shows that $\E[\|\M^{\star}\|^{2}]\geq \E[\|\M\|^{2}]=n$. This means that the second construction increases the variance of the score under $\mathcal{H}_{0}$. 

To make further progress, we resort to the asymptotic theory of random matrices, and especially to the Marcenko-Pastur distribution. Suppose that both $d$ and $n$ tend to infinity while remaining proportional: $n=cd$ with $c<1$. The matrix $\X^{\top}\X$ can be thought of as an empirical covariance matrix of the $d$ vectors which are the rows of $\X$. These vectors are i.i.d.~with bounded support components. Therefore, the marginal distribution of the $n$ eigenvalues $\{\lambda_{i}\}_{i=1}^{n}$ of $\X^{\top}\X$ asymptotically follows the Marcenko-Pastur distribution: for all $\lambda$ such that $(1-\sqrt{c})^{2} \leq \lambda \leq (1+\sqrt{c})^{2}$,
\begin{align}
f_\mathrm{MP}(\lambda)=\frac{\sqrt{(\lambda-(1-\sqrt{c})^{2})((1+\sqrt{c})^{2}-\lambda)}}{2c\pi\lambda}.
\end{align}
Moreover, for any function $\psi$ bounded over the interval $[(1-\sqrt{c})^{2},(1+\sqrt{c})^{2}]$:
\begin{equation}
\frac{1}{n}\sum_{i=1}^{n}\psi(\lambda_{i}) - \int \psi(\lambda)f_\mathrm{MP}(\lambda)d\lambda \rightarrow 0.
\end{equation}

Using the eigendecomposition, $\X^{\top}\X = \U\Lambda\U^{\top}$, we get $(\X^{\top}\X)^{-1} = \U\Lambda^{-1}\U^{\top}$, where the columns of $\U$ are the random eigenvectors. Then, asymptotically as $d\rightarrow\infty$:
\begin{align}
n^{-1}\E[\mathbf{1}_{n}^{\top}(\X^{\top}\X)^{-1}\mathbf{1}_{n}]-\int \lambda^{-1}f_\mathrm{MP}(\lambda)d\lambda \rightarrow 0.
\end{align}
This shows that $\E[\|\M^{\star}\|^{2}]/n$ converges to $\frac{1}{1-c}$ asymptotically (the above integral is the Stieljes transform of the Marcenko-Pastur distribution evaluated at $z=0$), whereas $\E[\|\M\|^{2}]/n$ converges to $\E[\lambda]=1$. In expectation, $\M^{\star}$ has a higher variance, but only by a factor of $1/(1-c)$ which remains acceptable if $c=n/d$ is small.

\subsection{$\Y$ uniformly drawn over a spherical cap}
\label{App:SphericalCap}
Assume that $\Y$ is uniformly distributed over the spherical cap $\mathcal{C}_{\u,\gamma}$, which is the intersection of the unit hypersphere and the single hypercone of axis $\u$, $\|\u\|=1$ and angle $\gamma$. In other words, $\|\Y\|=1$ and $S' = \Y^{\top}\u>\cos(\gamma)$. Denote $\eta = \cos(\gamma)$ and $\bar{\eta} = \mathsf{sign}(\eta)$. The probability distribution function of $S'$ is
\begin{equation}
f_{S'}(s') = \frac{f_{S}(s')}{1-F_{S}(s')}\mathbb{1}_{[s'>\eta]}(s').
\end{equation}
This stems into
\begin{equation}
\label{eq:ExpSprime}
\E[S'] = \frac{2(1-\eta^{2})^{\frac{d-1}{2}}}{(d-1)B(\frac{1}{2},\frac{d-1}{2})\left(1-\bar{\eta}I_{\eta^{2}}(\frac{1}{2},\frac{d-1}{2})\right)}.
\end{equation}
Note that:
\begin{itemize}
\item $\eta=-1$: $\E[S']=0$. The cap is the full hypersphere.
\item $\eta\rightarrow1$: $\E[S']\rightarrow 1$, thanks to De l'Hospital's rule. The cap reduces to $\{\u\}$.
\end{itemize}
In the same way:
\begin{equation}
\label{eq:ExpSprime2}
\E[S'^{2}] = \frac{1}{d}\frac{1-\bar{\eta}I_{\eta^{2}}(\frac{3}{2},\frac{d-1}{2})}{1-\bar{\eta}I_{\eta^{2}}(\frac{1}{2},\frac{d-1}{2})},
\end{equation}
from which we can deduce $\V(S')$ with the K¬\"onig-Huygens formula.
Note that:
\begin{itemize}
\item $\eta=-1$: $\V[S']=1/d$.
\item $\eta\rightarrow1$: $\V[S']\rightarrow 0$, thanks to De l'Hospital's rule.
\end{itemize}
From now on, we define
\begin{equation}
\label{eq:Mu}
\mu_{\kappa}(\eta,d) = \E[(S')^{\kappa}],
\end{equation}
with $S'=\Y^{\top}\u$ and $\Y\sim\mathcal{U}_{\mathcal{C}_{\u,\gamma}}$.

\subsection{Modeling k-means for the sum construction}
\label{sec:ModelingKmeanSum}
We assume that $k$-means has packed together in a memory unit independent vectors uniformly distributed over the spherical cap: $\x_{i}\sim\mathcal{U}_{\mathcal{C}_{\u,\gamma}}$.
Vector $\x_{i}$ can be modeled as $\x_{i} = S'_{i}\u+\N_{i}$, where $\{S'_{i}\}_{i=1}^{n}$ are i.i.d. according to the pdf described in Section~\ref{App:SphericalCap}, $\N_{i}^{\top}\u=0$ and $\|\N_{i}\|^{2}=1-(S^{\prime}_{i})^{2}$. 
Their memory vector is the sum $\m=\sum_{i=1}^{n}\x_{i}$.
\subsubsection{Hypothesis $\mathcal{H}_{0}$}
$\Y$ is independent from $\m$. It follows that:
\begin{equation}
\label{eq:SumRAH0}
\E[\Y^{\top}\m] = 0,\quad \V[\Y^{\top}\m] = \E[\|\m\|^{2}]/d
\end{equation}
with
\begin{eqnarray}
\E[\|\m\|^{2}]&=&\E[\sum_{i=1}^{n}\|\x_{i}\|^{2}+2\sum_{i<j}S'_{i}S'_{j}+\N_{i}^{\top}\N_{j}]\\
&=& n + n(n-1)\mu_{1}(\eta,d)^{2},
\end{eqnarray}
where $\mu_{1}(\eta,d)$ is given in~\eqref{eq:ExpSprime}.

To sum up: while $\E[\Y^{\top}\m]=0$, the variance is increasing with $\eta$.
\begin{itemize}
\item $\eta=-1$: It equals $n/d$ as already shown by~\eqref{eq:ModelH0Sum}.
\item $\eta\rightarrow1$: It converges to $n^{2}/d$.
\end{itemize}

\subsubsection{Hypothesis $\mathcal{H}_{1}$}
\label{sec:H1Sum}
We now single out the role of the matching vector $\X_{1}$:
$\m = \x_{1} + \m'$. The query is modeled as $\Y=\alpha\x_{1}+\beta\Z$ with $\|\Z\|=1$ and $\Z^{\top}\x_{1}=0$. With the same notation as above:
\begin{equation}
\Y^{\top}\m = \alpha + \alpha \sum_{i=2}^{n}S_{1}S_{i}+\alpha \N_{1}^{\top}\m'+\beta\Z^{\top}\m'.
\end{equation}
The expectation easily comes as
\begin{equation}
\E[\Y^{\top}\m] = \alpha(1+ (n-1)\mu_{1}(\eta,d)^{2}).
\end{equation}
The variance of the second summand is given by the law of total variance:
\begin{equation}
\V[\alpha \sum_{i=2}^{n}S_{1}S_{i}] = \alpha^{2}(n-1)\left(\mu_{2}(\eta,d)^{2}-\mu_{1}(\eta,d)^{4}\right).
\end{equation}
The variance of the third term is
\begin{equation}
\V[\alpha \sum_{i=2}^{n}\N_{1}^{\top}\N_{i}] = \frac{\alpha^{2}(n-1)}{d-1}(1-\mu_{2}(\eta,d))^{2}.
\end{equation}
The variance of the last summand is more complex to analyze. We need to decompose $\Z$ into its projection on $\u$ and on the complementary space. 
\begin{eqnarray*}
\V[\beta (\Z^{\top}\u)\u^{\top}\m']&=&(1-\alpha^{2})\frac{n-1}{d-1}(1-\mu_{2}(\eta,d))\mu_{2}(\eta,d)\\
\V[\beta \Z_{\bot}^{\top}\m']&=&(1-\alpha^{2})\frac{n-1}{d-1}(1-\mu_{2}(\eta,d))\\
&&\times (1+(n-2)\mu_{1}(\eta,d)^{2})
\end{eqnarray*}


To sum up: $\E[\Y^{\top}\m]$ increases while $\V[\Y^{\top}\m]$ decreases with $\eta$.
\begin{itemize}
\item $\eta=-1$: $\E[\Y^{\top}\m]=\alpha$ while $\V[\Y^{\top}\m] = (n-1)/d$ as already shown by~\eqref{eq:ModelH1Sum}.
\item $\eta\rightarrow1$: $\E[\Y^{\top}\m]\rightarrow n\alpha$ whereas $\V[\Y^{\top}\m]\rightarrow 0$.
\end{itemize}

\begin{figure}
        \includegraphics[width=0.5\textwidth]{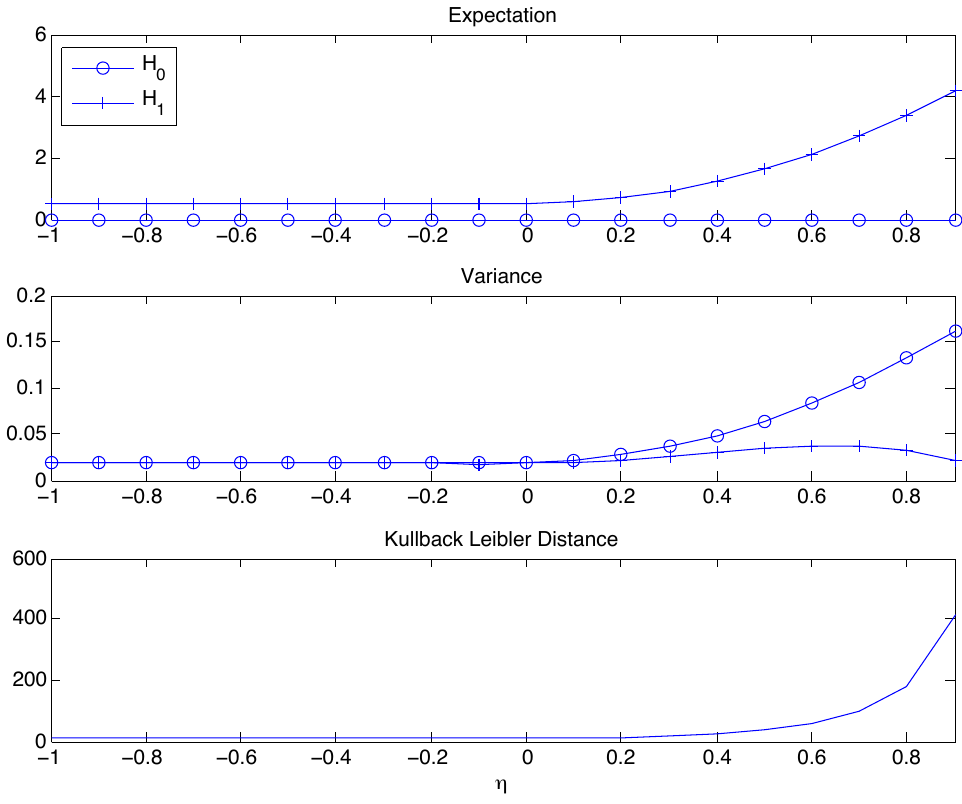}
\caption{Expectations and Variances under both hypothesis and Kullback Leibler distance as functions of $\eta$ for the sum construction. $\alpha=0.5$, $d=512$.
\label{fig:H0_app}}
\end{figure}

In the end, under the Gaussian assumption, the Kullback-Leibler distance between both distributions increases which proves that identifying the positive memory units becomes easier as $\eta$ increases (see Fig.~\ref{fig:H0_app}). 

\subsection{Modeling k-means for the pinv construction}
\label{sec:ModelingKmeanPinv}

\subsubsection{Hypothesis $\mathcal{H}_{0}$}
We make the same assumption as in Section~\ref{sec:ModelingKmeanSum}.
We write $\X$ as $\X = \u\S'^{\top}+\N$ with $\S'$ a $n\times 1$ vector storing the correlations $\u^{\top}\x_{i}$, $1\leq i\leq n$ and $\N$ a $d\times n$ matrix whose columns are random vectors orthogonal to $\u$ and of norm $\sqrt{1-S^{\prime 2}_{i}}$. The memory unit is now given by~\eqref{equ:pinv}: $\m^{\star}=\X(\X^{\top}\X)^{-1}\1_{n}$.
Eq.~\eqref{eq:SumRAH0} holds but with a new expression for the norm of $\m^{\star}$:
\begin{eqnarray}
\E[\|\m^{\star}\|^{2}]&=&\E[\1_{n}^{\top}(\X^{\top}\X)^{-1}\1_{n}]\\
&\geq&  \1_{n}^{\top}(\E[\X^{\top}\X])^{-1}\1_{n}.
\end{eqnarray}
We write matrix $\X^{\top}\X=\X^{\top}\X = \S'\S'^{\top} + \N\N^{\top}$ s.t.:
\begin{equation}
\E[\X^{\top}\X] = (1-\E[S']^{2})\mathbf{I}_{n} + \E[S']^{2} \1_{n}\1_{n}^{\top},
\end{equation}
whose inverse is given by the Sherman-Morrison formula:
\begin{equation*}
\E[\X^{\top}\X]^{-1} = \frac{1}{1-\E[S']^{2}}\left(\mathbf{I}_{n} + \frac{\E[S']^{2}}{1+(n-1)\E[S']^{2}} \1_{n}\1_{n}^{\top}\right),
\end{equation*}
leading to
\begin{equation}
\E[\|\m^{\star}\|^{2}] \geq \frac{n}{1+(n-1)\E[S']^{2}}
\end{equation}

To sum up: While $\E[\Y^{\top}\m^{\star}]$ remains constant, the lower bound of the variance is decreasing with $\eta$.
\begin{itemize}
\item $\eta=-1$: The lower bound equals $\frac{n}{d}$ which is tight w.r.t. to the previous result in Sect.~\ref{sec:AppExp} (i.e. $\frac{n}{d}(1+\frac{n}{(d-n)})$) for $n$ much smaller than $d$.
\item $\eta\rightarrow1$: The lower bounds converges to $1/d$ and is also tight since $\m^{\star}\rightarrow \x_{1}$ as the vectors are converging to $\x_{1}$.
\end{itemize}

\subsubsection{Hypothesis $\mathcal{H}_{1}$}
We single out the matching vector $\x_{1}$ by writing $\X = (\x_{1},\bar{\X})$  with $\bar{\X} = \u\S'^{\top}+\N$, $\S'$ being now a $(n-1)\times 1$ vector storing the correlations $\u^{\top}\x_{i}$, $2\leq i\leq n$ and $\N$ a $d\times (n-1)$ matrix whose columns are random vectors orthogonal to $\u$ and of norm $\sqrt{1-S^{\prime 2}_{i}}$.
This makes
\begin{equation}
\X^{\top}\X = \left(\begin{array}{cc}1 & \x_{1}^{\top}\bar{\X} \\ \bar{\X}^{\top}\x_{1}& \bar{\X}^{\top}\bar{\X}\end{array}\right),
\end{equation}
whose inverse is
\begin{equation}
(\X^{\top}\X)^{-1} = \left(\begin{array}{cc}1+\x_{1}^{\top}\bar{\X}\D\bar{\X}^{\top}\x_{1} & - \x_{1}^{\top}\bar{\X}\D \\-\D\bar{\X}^{\top}\x_{1} & \D\end{array}\right).
\end{equation}
with $\D = (\bar{\X}^{\top}(\I-\X_{1}\X_{1}^{\top})\bar{\X})^{-1}$.
This makes the following memory vector:
\begin{eqnarray}
\m^{\star}&=&\x_{1} + \m^{\star}_{\bot}\\
\m^{\star}_{\bot}&=&(\I-\x_{1}\x_{1}^{\top})\bar{\X}\D(\1_{n-1}-\bar{\X}\x_{1})
\end{eqnarray}
Note that $\|\m^{\star}\|^{2} = 1 + \|\m^{\star}_{\bot}\|^{2}$ because $\x_{1}^{\top}\m^{\star}_{\bot}=0$.

%
The query vector being defined as in Sect.~\ref{sec:H1Sum}, its correlation with the memory vector is
\begin{equation}
\Y^{\top}\m^{\star} = \alpha + \sqrt{1-\alpha^{2}}\Z^{\top}\m^{\star}_{\bot},
\end{equation}
whose expectation and variance are given by
\begin{equation}
\E[\Y^{\top}\m^{\star}]=\alpha,\quad \V[\Y^{\top}\m^{\star}] = (1-\alpha^{2})\frac{\E[\|\m^{\star}_{\bot}\|^{2}]}{d-1}
\end{equation}
with
\begin{eqnarray}
\E[\|\m^{\star}_{\bot}\|^{2}]&=&\E[\|\m^{\star}\|^{2}]-1\\
&\geq& \frac{(n-1)(1-\E[S']^{2})}{1+(n-1)\E[S']^{2}}.
\end{eqnarray}

To sum up: While $\E[\Y^{\top}\m^{\star}]$ remains constant, the lower bound of the variance is decreasing with $\eta$.
\begin{itemize}
\item $\eta=-1$: The lower bound equals $(1-\alpha^{2})\frac{n-1}{d-1}$ which is tight w.r.t. to the previous result in Sect.~\ref{sec:AppExp} (i.e. $(1-\alpha^{2})\frac{n}{d-n}$) for $n$ much smaller than $d$.
\item $\eta\rightarrow1$: The lower bounds converges to $0$ and is also tight since $\m^{\star}\rightarrow \x_{1}$ as the vectors are converging to $\x_{1}$.
\end{itemize}

\begin{figure}
        \includegraphics[width=0.5\textwidth]{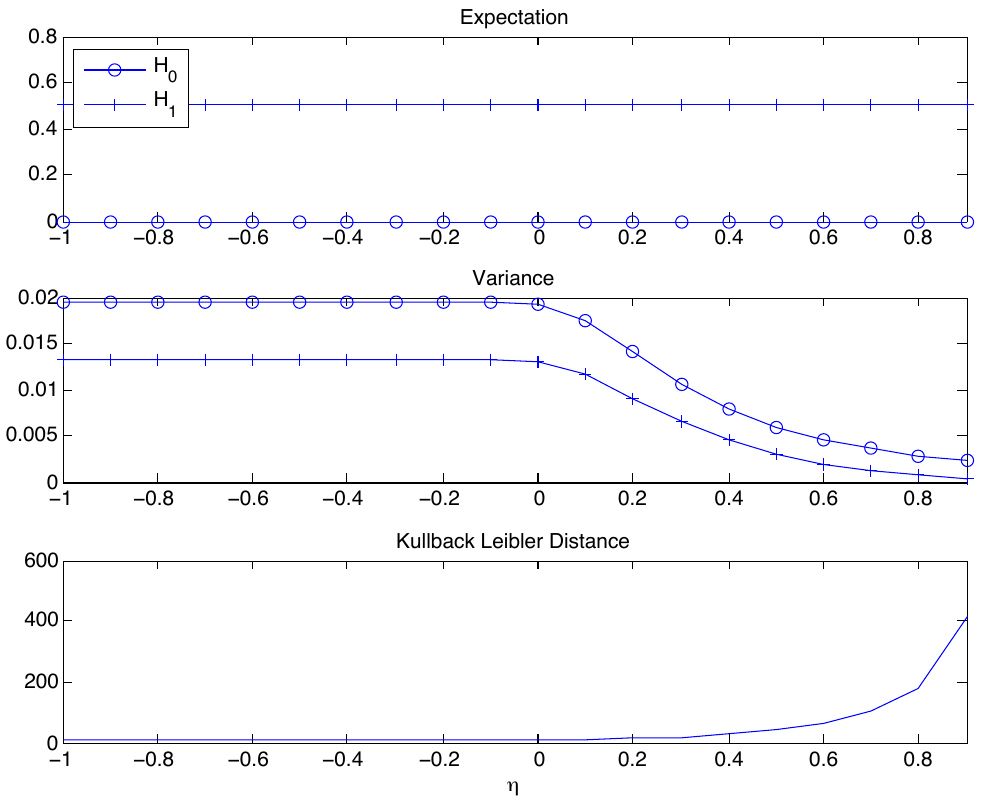}
\caption{Expectations and Variances under both hypothesis and Kullback Leibler distance as functions of $\eta$ for the pinv construction. $\alpha=0.5$, $d=512$.
\label{fig:H1_app}}
\end{figure}

In the end, under the Gaussian assumption, the Kullback-Leibler distance between both distributions increases which proves that identifying the positive memory units becomes easier as $\eta$ increases (see Fig.~\ref{fig:H1_app}). 

